\documentclass[10pt,twocolumn,letterpaper]{article}

\usepackage[pagenumbers]{iccv} % To force page numbers, e.g. for an arXiv version

\definecolor{iccvblue}{rgb}{0.21,0.49,0.74}
\usepackage[pagebackref,breaklinks,colorlinks,allcolors=iccvblue]{hyperref}
\usepackage{multirow}
\usepackage{xcolor}
\usepackage{colortbl}
\usepackage{amssymb}
\usepackage{url}
\def\paperID{***} % *** Enter the Paper ID here
\def\confName{ICCV}
\def\confYear{2025}

\title{
Video-LevelGauge: Investigating Contextual Positional Bias \\ in Large Video Language Models
}

\author{
Hou Xia$^{1}$, Zheren Fu$^{1}$, Fangcan Ling$^{1}$, Jiajun Li$^{2}$, Yi Tu$^{2}$, Zhendong Mao$^{1}$\thanks{Corresponding Author.}, Yongdong Zhang$^{1}$ \\
$^{1}$University of Science and Technology of China, Hefei, China\\
$^{2}$HUAWEI, Shanghai, China \\
{\tt\small \{overwhelmed, fzr, lfc200250\}@mail.ustc.edu.cn, \{zdmao, zhyd73\}@ustc.edu.cn} \\
{\tt\small \{jiajun.work, xssg.tuyi\}@huawei.com
}
}

\begin{document}
\maketitle
\begin{abstract}
Large video language models (LVLMs) have made notable progress in video understanding, spurring the development of corresponding evaluation benchmarks. However, existing benchmarks generally assess overall performance across entire video sequences, overlooking nuanced behaviors such as contextual positional bias, a critical yet under-explored aspect of LVLM performance. 
We present \textbf{Video-LevelGauge}, a dedicated benchmark designed to systematically assess positional bias in LVLMs. We employ standardized probes and customized contextual setups, allowing flexible control over context length, probe position, and contextual types to simulate diverse real-world scenarios. In addition, we introduce a comprehensive analysis method that combines statistical measures with morphological pattern recognition to characterize bias.
Our benchmark comprises 438 manually curated videos spanning multiple types, yielding 1,177 high-quality multiple-choice questions and 120 open-ended questions, validated for their effectiveness in exposing positional bias. 
Based on these, we evaluate 27 state-of-the-art LVLMs, including both commercial and open-source models.
Our findings reveal significant positional biases in many leading open-source models, typically exhibiting head or neighbor-content preferences. In contrast, commercial models such as Gemini2.5-Pro show impressive, consistent performance across entire video sequences. Further analyses on context length, context variation, and model scale provide actionable insights for mitigating bias and guiding model enhancement.
\href{https://github.com/Cola-any/Video-LevelGauge}{https://github.com/Cola-any/Video-LevelGauge}.
\end{abstract}    
\section{Introduction}
\label{sec:intro}
Large Video Language Models (LVLMs) have advanced rapidly in recent years, revolutionizing video understanding by integrating large language models (LLMs) with visual perception capabilities. These models~\cite{llava-video, videollama3, gpt-4o} show impressive performance across diverse video tasks, with strong generalization abilities. Alongside these advancements, continuous efforts have been devoted to developing video-centric benchmarks to assess their effectiveness, providing critical insights for future improvement.

\begin{figure}[t]
\centering
\includegraphics[width=0.99\columnwidth]{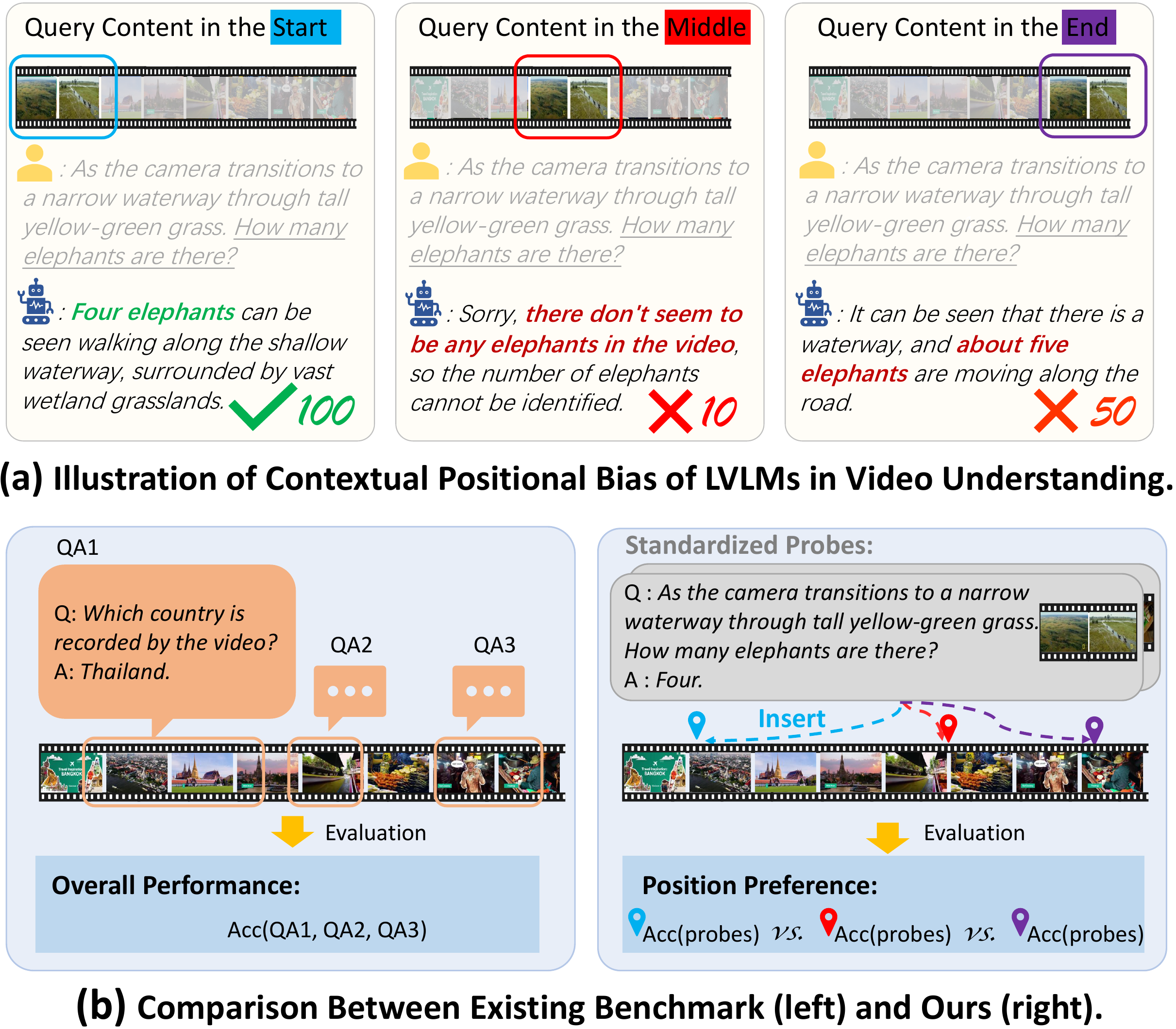} 
\caption{
(a) Large Video Language Models (LVLMs) suffer from positional bias, characterized by uneven comprehension of identical content presented at different contextual positions.
(b) Existing benchmarks typically assess models based on overall performance across the entire video sequence, which is limited in revealing this nuanced behavior. Video-LevelGauge explicitly investigates this by inserting standardized probes (curated segments tagged with meticulous questions) into varying positions of the context.
} 
\label{fig1}
\end{figure}

Existing video benchmarks~\cite{video-chatgpt,tempcompass} primarily evaluate models based on overall performance across entire video sequences on a variety of tasks, such as temporal reasoning and summarization. Notable benchmarks include MVBench~\cite{mvbench}, TempCompass~\cite{tempcompass}, and MMVU~\cite{mmvu} for short videos, as well as MLVU~\cite{MLVU}, LongVideoBench~\cite{longvideobench}, and VideoMME~\cite{videomme} for long video understanding.
While these benchmarks offer broad task assessments, they provide limited insight into how models interpret content at different contextual positions. As illustrated in \cref{fig1} (a), LVLMs can suffer from positional bias, i.e., inconsistent comprehension of identical content presented at varying contextual locations. However, existing benchmarks can offer minimal diagnostic value for mitigating this issue.

\begin{figure}[t]
\centering
\includegraphics[width=0.99\columnwidth]{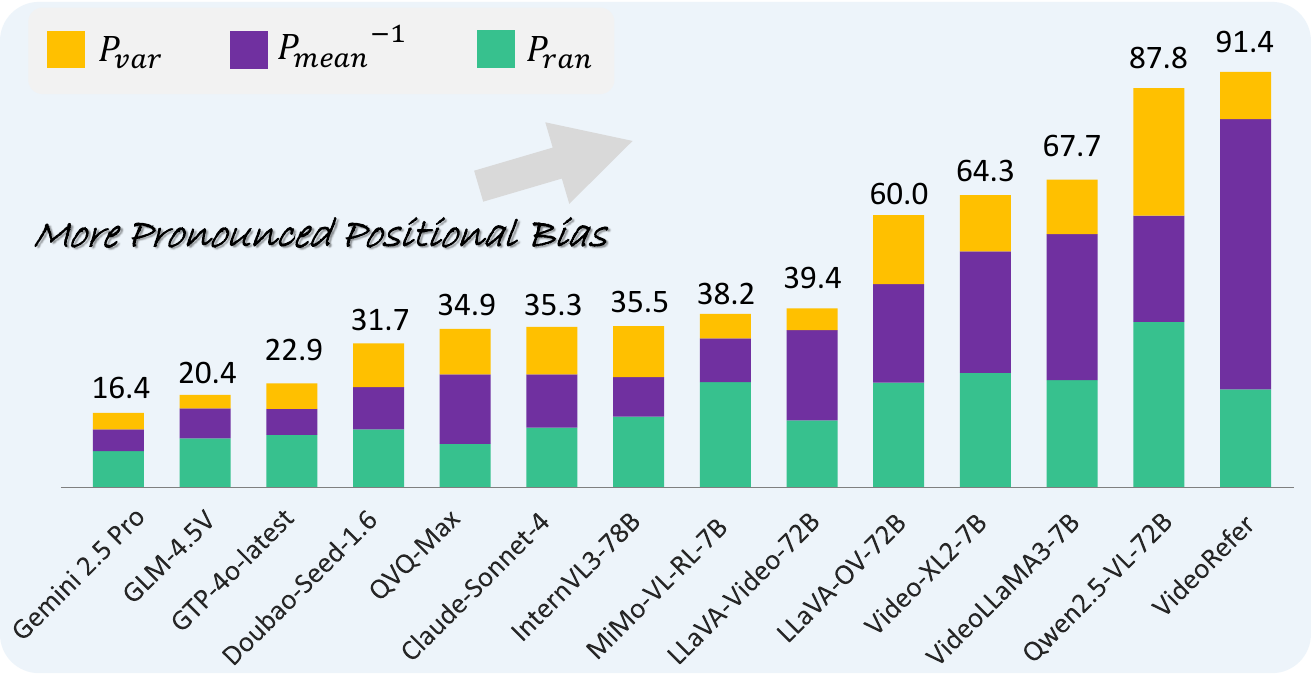} 
\caption{
\textbf{Performance of state-of-the-art LVLMs on Video-LevelGauge}, evaluated by our composite metric, where higher values indicate more pronounced positional bias. Gemini 2.5 Pro~\cite{gemini25} exhibits the least positional bias, followed by GLM-4.5V~\cite{glm45v}, GPT-4o-latest~\cite{gpt-4o}, Doubao-Seed-1.6~\cite{doubao-seed}, and others.
} 
\label{fig:leaderborad}
\end{figure}

The serial position effect~\cite{serialeffect} in psychology suggests that humans tend to better recall content presented at the beginning and end of a sequence. Similar behaviors have been observed in language models~\cite{lostmiddle}. To date, how various types of LVLMs, such as those incorporating memory components~\cite{Ma-LMM} or trained with long-context~\cite{video-xl2}, perform on positional biases remains under-explored.
Moreover, how positional bias manifests in video–text interleaved contexts is still an open question. In particular, models claiming to excel at long video understanding should be validated for their ability to maintain consistent and effective perception across the entire sequence, with minimal positional bias.

To this end, we introduce \textbf{Video-LevelGauge}, a dedicated benchmark for evaluating contextual positional bias in video understanding. Inspired by needle-in-a-haystack approaches~\cite{needlehay, NIAVH}, we propose a standardized probe and customized context strategy for constructing evaluation data. As shown in \cref{fig1} (b), our benchmark inserts standardized probes at varying positions within the context to assess models’ bias to contextual positions.
It supports flexible control over context length, probe position, and context composition to evaluate positional biases in various real-world scenarios, such as long video comprehension and multi-modal interleaved inputs.
In addition, we propose a comprehensive analysis method for positional bias that combines statistical metrics and morphological recognition. 
Video-LevelGauge includes 438 manually curated multi-type videos and 1,177 high-quality multiple-choice question answering items spanning six tasks, along with 120 open-ended descriptive questions, constructed through a labor-efficient workflow. These questions are empirically validated to be highly sensitive to visual perception, making them well-suited for detecting positional bias.
 
We evaluate 27 state-of-the-art LVLMs, including 6 commercial and 21 open-source models based on diverse techniques. We reveal that most leading open-source models suffer from positional bias. They manifest various bias patterns, such as neighbor preference, U-shaped curves, and head preference. 
In comparison, commercial models, e.g., Gemini 2.5 Pro, achieve consistent and superior performance across the entire sequence, as presented in \cref{fig:leaderborad}. 
This indicates substantial room for improvement in mitigating positional bias among open-source models, which could significantly enhance their understanding capacity. 
Further analysis of context length, context variation, and model scale provides three key findings for positional bias in LVLMs.
Contributions are summarized as follows:

\begin{itemize}
    \item We emphasize the necessity of positional bias evaluation as a complement to existing benchmarks. To the best of our knowledge, this paper is the first systematic study of contextual positional bias in video understanding.
    \item We propose Video-LevelGauge, a tailored benchmark for evaluating positional bias in video understanding with a comprehensive analysis method, which supports flexible configurations over context lengths and probe positions.
    \item We experimentally investigate positional bias in 27 state-of-the-art LVLMs and conduct an in-depth analysis of the effects of context length, context variation, and model size, offering insights for future refinement.
\end{itemize}

\section{Related Work}
\label{sec:related work}

\subsection{Large Video Language Model (LVLM)}
LVLMs have made significant strides with the integration of large language models and visual encoders. 
Early works~\cite{video-llava,video-chatgpt} focus on short video understanding through video post-training. Later efforts~\cite{kangaroo, llama-vid} extend to long video understanding. Two-stage methods~\cite{llovi, lvnet} perform video understanding by inputting frame descriptions into LLMs. Others~\cite{movie-chat, Ma-LMM} incorporate explicit memory modules to capture long-term dependencies. 
\begin{figure*}[t]
\centering
\includegraphics[width=0.99\textwidth]{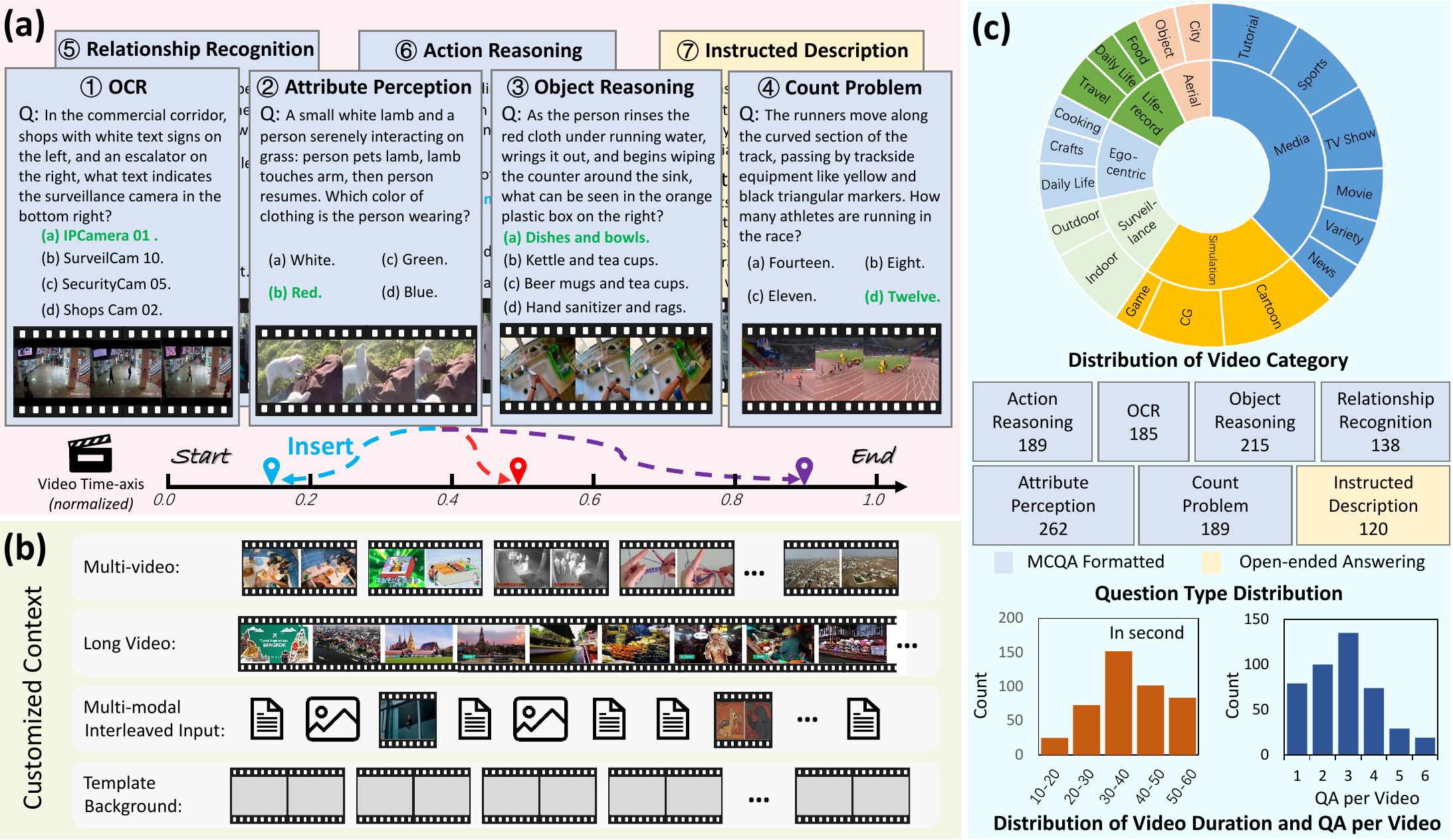} 
\caption{
\textbf{Overview of Video-LevelGauge}, our benchmark for contextual positional bias in video understanding. It adopts a standardized probe and customized context paradigm, where crafted probes are inserted at varying positions within context.
(a) Examples of standardized probes on six multi-choice question answering (MCQA) formatted evaluation tasks and one open-ended instructed description task.
(b) Four customized context types for investigating positional bias under various real-world scenarios.
(c) Benchmark statistics. 438 multi-type videos, 1,177 high-quality MCQA problems, and 120 open-ended items are constructed.
}
\label{fig2}
\end{figure*}
Besides, models such as LongVA~\cite{longva}, LongVILA~\cite{longvila} and Video-XL2~\cite{video-xl2} extend contextual lengths through long video training.
More recently, models like Qwen2.5-VL~\cite{qwen25vl}, InternVL3~\cite{internvl3}, and LLaVA-OV~\cite{llavaonevision} built upon long-context LLMs, exhibit strong performance in long video tasks. MiMo-VL~\cite{mimovl} and GLM-4.5V~\cite{glm45v} deliver impressive performance in video understanding equipped with multimodal reasoning.
Task-specific models have also been developed. VideoRefer~\cite{videorefer} enhances fine-grained spatial-temporal understanding and T-Star~\cite{T-star} improves contextual search. 
In this paper, we investigate whether these models can effectively comprehend entire videos without positional bias.

\subsection{Video Benchmark}
Alongside the advancement of LVLMs, many efforts have been made to develop video-centric benchmarks.
Some benchmarks~\cite{mvbench, tempcompass} focus on evaluating short-video understanding. To extend evaluation to long-videos, VideoMME~\cite{videomme}, MLVU~\cite{MLVU}, LongVideoBench~\cite{longvideobench}, LVBench~\cite{lvbench}, and HLV-1K~\cite{hlv1k} focus on constructing videos of long duration and designing tasks involving long-range comprehension. 
Another line of work adopts synthetic construction strategies, named Needle-in-a-Haystack (NIAH). For instance, some works~\cite{lv-eval, needlehay} target long-text comprehension for LLMs, while others~\cite{multimodalneedlehaystack, needlemultimodalhaystack} focus on multi-modal contexts. 
More recently, VNBench~\cite{NIAVH}, LV-Haystack~\cite{T-star}, and V-NIAH~\cite{longva} are proposed to assess temporal search capabilities in LVLMs. 
Unlike these benchmarks, which primarily assess models' overall performance across the entire video sequence, our benchmark is specifically designed to evaluate the contextual positional bias in video understanding, providing a dedicated analysis method to reveal nuanced model behaviors.
\section{Video-LevelGauge}
\label{sec:method}

We begin with an overview, emphasizing our design philosophy and statistics (\cref{sec:3.1}), followed by descriptions of probe QA construction (\cref{sec:3.2}), context construction (\cref{sec:3.3}), and our analysis method (\cref{sec:3.4}). 

\subsection{Overview}
\label{sec:3.1}
As shown in \cref{fig2}, Video-LevelGauge introduces a standardized probe and customized context design paradigm, where carefully designed probe segments are inserted at varying positions within customized contextual contents. By comparing model responses to identical probes at different insertion points, we assess positional bias in video comprehension. Compared to approaches~\cite{MLVU,hlv1k} that densely formulate QA pairs in natural videos, this offers three advantages. (1) Controlled variables, eliminating confounding effects of varying QA difficulty at different positions. (2) Flexible control over both the context length and evaluated positions. (3) Simulation of diverse real-world scenarios, such as multi-video understanding, long videos, and interleaved video-text inputs.

Video-LevelGauge encompasses six categories of structured video understanding tasks (e.g., action reasoning), along with an open-ended descriptive task. It includes 438 manually collected multi-type videos, 1,177 multiple-choice question answering (MCQA) items, and 120 open-ended instructed descriptive problems paired with annotations. Each question is described with scene mention and task instruction to ensure clarity, requiring genuine visual comprehension. In this way, Video-LevelGauge provides a comprehensive evaluation of positional bias in LVLMs.

\subsection{Standardized Probe Construction}
\label{sec:3.2}
Probe construction is carried out in three steps: video collection, QA generation, and probe requirement validation.

\subsubsection{Video Collection.}
As shown in \cref{fig2} (c), six types of videos are collected from public test sets to avoid data leakage and ethical concerns. Specifically, it includes 42 aerial videos from VisDrone~\cite{visdrone}, 49 surveillance videos from UCF-Crime~\cite{ucf-crime}, 50 egocentric videos from Ego-4D~\cite{ego4d}, 152 media videos, 58 life-record videos, and 87 synthetic videos from MLVU and VideoMME. PySceneDetect is used to segment the original videos, followed by manual selection. 
Blurry, static, and duplicate videos are filtered out.

\begin{figure}[t]
\centering
\includegraphics[width=0.99\columnwidth]{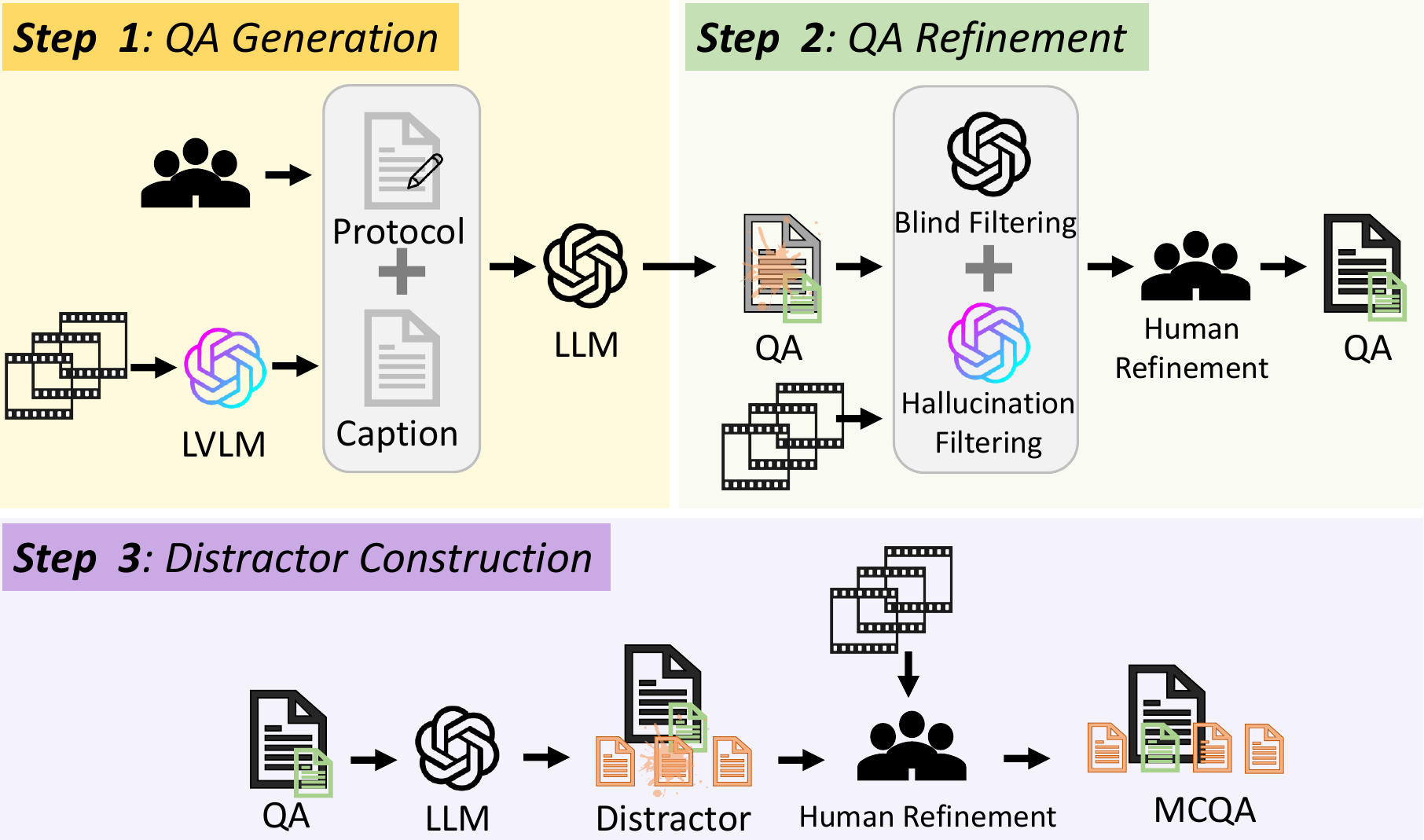} 
\caption{\textbf{Probe QA construction workflow.} We propose a labor-efficient workflow that combines automated generation with human refinement. Candidate QAs are first generated using LLM guided by manually defined protocols. These QAs are then enhanced using both LLM and LVLM to eliminate hallucinations and information leakage, followed by human refinement. Finally, for challenge MCQAs, distractor options are generated by LLM and double checked by annotators.
}
\label{fig3}
\end{figure}

\subsubsection{Construction of Multi-Task QA.}
To comprehensively evaluate positional bias across various video understanding tasks, we construct six types of structured MCQA tasks: Optical Character Recognition (OCR), Attribute Perception (AP), Object Reasoning (OR), Counting Problem (CP), Relationship Recognition (RR), and Action Reasoning (AR), along with one open-ended descriptive task. 
As shown in \cref{fig3}, we construct probe QAs via a three-step workflow,
involving automated generation and human refinement.

\textit{(1) QA Generation}. Collected videos are frame-wise captioned using GPT-4o~\cite{gpt-4o}, and human annotators annotate the task definition along with three positive and negative QA examples for each video task. These video captions, task definitions, and QA examples are then fed into LLMs to generate task-specific question–answer pairs via prompt engineering.

\textit{(2) QA Refinement}. The generated QA pairs may be problematic. We perform blind filtering with LLMs to eliminate questions that leak answers or can be answered using commonsense, and use GPT-4o to filter out items with hallucinations or incorrect answers based on the original videos. Finally, all QA pairs undergo manual refinement.

\textit{(3) Distractor Construction}. For MCQAs, LLMs are prompted to generate multiple deceptive distractor choices, which are then manually vetted and refined by annotators to ensure clarity and no obvious difference among the options. The choices are randomly shuffled, and the average length of  questions is 30.4 words.

\begin{figure}[t]
\centering
\includegraphics[width=0.99\columnwidth]{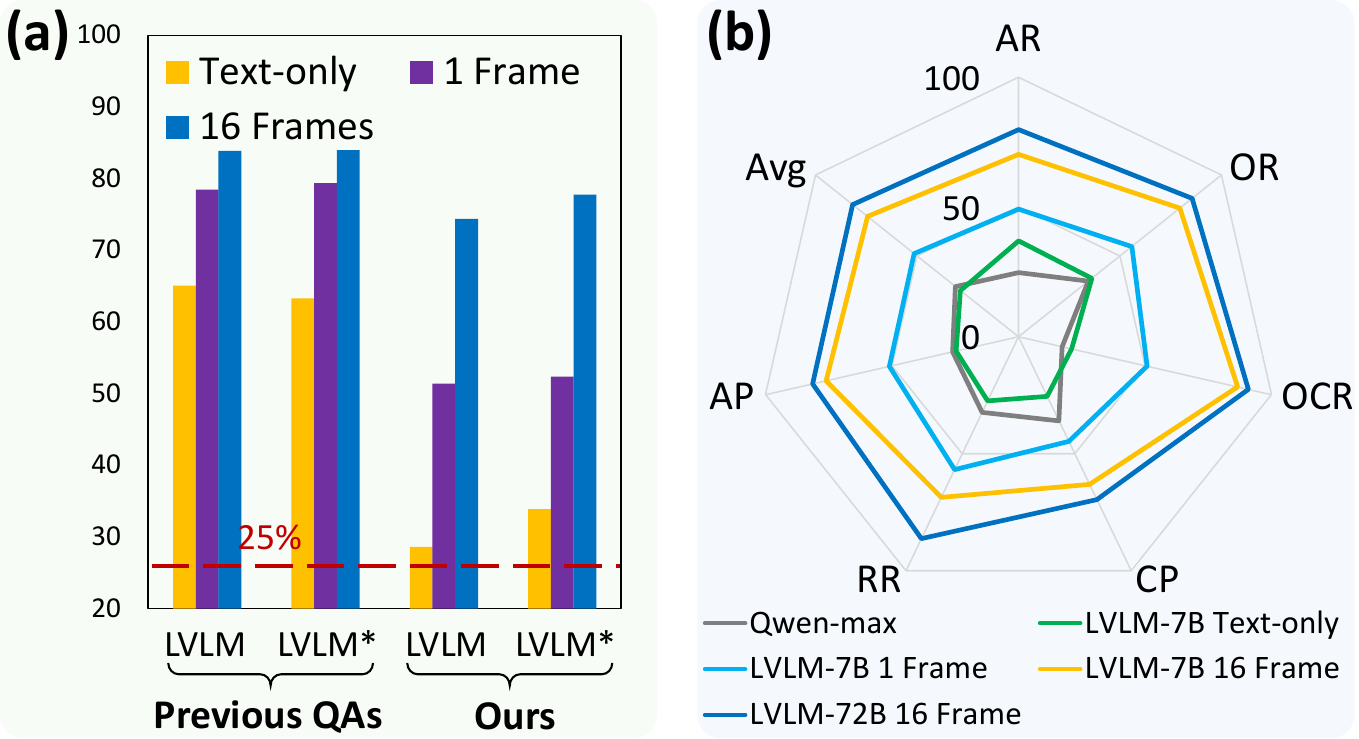} 
\caption{\textbf{Probe validation.} (a) Compared to previous needle QAs~\cite{MLVU}, our questions yield near-random accuracy when using text-only inputs on Qwen2.5-VL-7B (LVLM) and InternVL3-8B (LVLM*). 
This indicates that our QAs are highly sensitive to the degree of visual perception, making them well-suited for detecting positional bias.
(b) Accuracy improves with larger models and more frames, highlighting the challenging nature of our QAs.
}
\label{fig4}
\end{figure}

\textbf{Probe Requirement Validation.}
Although blind filtering is applied during QA construction, there may exist multi-modal leakage~\cite{mm-star}. Therefore, we validate the quality of our QAs using Qwen2.5-VL and InternVL3. As shown in \cref{fig4}, compared to the needle samples from MLVU, models achieve near-random performance under text-only input and perform far from saturation with single-frame inputs in our MCQAs. This indicates that models' performance on our QAs is highly sensitive to the adequacy of visual content perception, rendering it well-suited for evaluating positional bias. 
Furthermore, the performance of each task scales with both number of frames and model size, enabling discrimination between models of varying capacities.

\begin{table*}[t]
\caption{\textbf{Contextual positional bias analysis of various LVLMs on Video-LevelGauge}. 
\textit{OCR, AP, OR, CP, RR}, and \textit{AR} represent six video understanding tasks respectively, as shown in \cref{sec:3.2}. 
\textit{Average} denotes the average performance of all six tasks. 
$P_{mean}$: the mean score across all evaluated positions; 
$P_{ran}$: the range of score between positions; 
$P_{var}$: the position variance; 
\textit{MR}: morphological recognition of bias pattern; 
$S_{meta}$: the accuracy with only probe input. 
Metric details in \cref{sec:3.4}. $^\dagger$ represents multi-modal reasoning.
}
\label{tab:main}
\resizebox{1.0\textwidth}{!}{
\begin{tabular}{l|c|cccccc|ccccc}
\hline
\noalign{\vskip 2pt}
\multirow{2}{*}{Models} & \multirow{2}{*}{Model size} & OCR&AP&OR&CP&RR&\multicolumn{1}{c|}{AR}  & \multicolumn{5}{c}{Average}  \\
 & & $P_{ran}\downarrow$& $P_{ran}\downarrow$& $P_{ran}\downarrow$ & $P_{ran}\downarrow$ & $P_{ran}\downarrow$ & $P_{ran}\downarrow$& $P_{mean}\uparrow$ & $P_{ran}\downarrow$ & $P_{var}\downarrow$ & \textit{MR} &$S_{meta}\uparrow$ \\ 
% \hline
\hline
\noalign{\vskip 2pt}
\multicolumn{13}{c}{\textit{Blind LLM}} \\
\noalign{\vskip 1pt}
\hline
\noalign{\vskip 2pt}
GPT-4~\cite{gpt4}& -  &-&-&-&-&-&-&-&- & - & - & 31.1  \\ 
\noalign{\vskip 1pt}
\hline
\noalign{\vskip 2pt}
\multicolumn{13}{c}{\textit{Two-stages Models}} \\
\noalign{\vskip 2pt}
\hline
\noalign{\vskip 2pt}
Caption + Qwen3~\cite{qwen3} & - &5.6&11.7&12.9&28.3&16.9&11.3&93.1&11.2&18.1&$\rm U$&65.7\\
Caption + GPT-4~\cite{gpt4}& - &\textbf{4.3}&\textbf{11.3}&\textbf{7.7}&\textbf{12.9}&\textbf{15.6}&\textbf{7.9}&\textbf{96.1}&\textbf{7.4}&\textbf{5.8}&$\rm U$&\textbf{67.7}\\ 
\noalign{\vskip 1pt}
\hline
\noalign{\vskip 2pt}
\multicolumn{13}{c}{\textit{Open-source Models}} \\
\noalign{\vskip 2pt}
\hline
\noalign{\vskip 2pt}
MA-LMM~\cite{Ma-LMM} & 7B &19.1&16.7&20.7&9.5&9.4&6.6&83.9&9.3&7.3&$\nearrow$&35.0\\
LongVA~\cite{longva} & 7B &9.6&21.6&15.8&7.7&17.1&9.1&82.4&9.2&7.8&$\rm U$&48.5\\
MiniGPT4-Video~\cite{minigpt4videos}&7B&26.9&21.9&14.9&10.4&5.7&\textbf{3.4}&84.9&9.6&15.4&$\searrow$&49.3\\
LLaMA-VID~\cite{llama-vid} & 13B &23.2&22.0&24.2&35.5&34.5&25.4&81.1&12.3&17.3&$\rm W$&31.2\\
Kangaroo~\cite{kangaroo} & 8B &14.3&13.2&10.4&24.6&38.9&15.6&89.2&8.2&8.4&$\rm W$&53.0\\
% LLaVA-OV~\cite{llavaonevision} & 7B &6.1&13.0&6.4&9.9&12.1&7.6&87.3&4.2&2.1&---&65.6\\
LLaVA-OV~\cite{llavaonevision} & 72B &\textbf{3.4}&7.0&5.4&\textbf{4.8}&15.5&6.7&92.8&5.8&3.8&$\nearrow$&72.0\\ 
LLaVA-Video~\cite{llava-video} & 7B &4.8&12.1&6.9&6.8&12.8&7.4&90.7&6.3&3.5&$\rm U$&70.0\\
LLaVA-Video~\cite{llava-video} & 72B &6.4&6.9&2.4&7.2&10.7&4.6&93.4&3.7&1.2&---&74.6\\
Qwen2.5-VL~\cite{qwen25vl} & 7B &8.1&27.4&9.8&16.9&31.0&15.0&89.6&12.4&11.7&$\rm U$&68.2\\
Qwen2.5-VL~\cite{qwen25vl} & 72B &5.6&20.2&15.8&22.2&17.7&7.3&92.2&9.1&7.0&$\searrow$&73.7\\
InternVL3~\cite{internvl3} & 8B &6.7&11.5&9.9&22.6&13.9&10.7&90.3&8.0&5.3&$\rm U$&70.5\\
InternVL3~\cite{internvl3} & 9B &9.0&9.1&5.5&20.3&13.4&10.3&95.7&\textbf{2.7}&1.7&---&67.3\\
InternVL3~\cite{internvl3} & 78B &4.7&10.6&5.0&21.4&14.7&4.5&97.1&3.9&2.8&---&74.2\\
VideoLLaMA2~\cite{videollama2} & 7B &10.7&6.2&10.3&18.2&15.8&14.1&91.4&5.6&6.1&$\rm U$&39.3\\
VideoLLaMA3~\cite{videollama3} & 7B &4.6&7.0&6.0&9.9&13.4&9.9&89.3&5.9&3.0&$\rm U$&76.2 \\
NVILA~\cite{NVILA} & 8B &12.5&34.2&17.2&15.1&17.7&14.0&79.1&13.9&14.7&$\nearrow$&64.5\\
LongVILA-1M~\cite{longvila} & 7B &7.3&22.4&14.4&10.7&12.7&12.4&81.6&11.5&12.7&$\rm U$&59.8\\
Video-XL~\cite{videoxl} & 7B &13.6&8.0&10.5&17.0&11.4&8.6&83.4&9.0&7.9&$\searrow$&47.7\\
Video-XL2~\cite{video-xl2} & 7B &\textbf{3.4}&13.2&6.8&13.2&7.2&6.9&91.1&6.3&3.1&$\rm U$&73.5\\
VideoRefer~\cite{videorefer} & 7B &4.6&9.6&7.5&8.4&11.4&7.6&80.2&5.4&2.6&$\searrow$&77.4\\
T-Star~\cite{T-star} & 7B &10.9&7.1&12.6&11.9&20.5&10.6&72.6&6.1&5.0&$\rm W$&69.4\\
MiMo-VL-RL~\cite{mimovl} & 7B &4.5&10.3&6.7&17.6&12.5&7.6&93.0&8.0&4.9&$\searrow$& 68.9 \\
MiMo-VL-RL$^\dagger$~\cite{mimovl} & 7B &5.4&6.4&6.5&21.5&6.9&9.0&96.8&5.8&1.8 &--- &69.6 \\
GLM-4.5V~\cite{glm45v} & 108B &4.1&\textbf{3.2}&2.8&8.0&6.0&8.0&97.2&3.4&1.4&---& 79.5 \\
GLM-4.5V$^\dagger$~\cite{glm45v} & 108B &\textbf{3.4}&5.2&\textbf{1.7}&6.6&\textbf{4.8}&6.7&\textbf{97.8}&\textbf{2.7}&\textbf{1.0}& --- & \textbf{79.9} \\
\noalign{\vskip 1pt}
\hline
\noalign{\vskip 2pt}
\multicolumn{13}{c}{\textit{Commercial Models}} \\
\noalign{\vskip 2pt}
\hline
\noalign{\vskip 2pt}
Qwen-VL-Max~\cite{qwenvl} & - &2.9&8.0&15.8&5.6&15.0&10.3&89.8&8.0&8.0&$\searrow$&70.5\\
QVQ-Max$^\dagger$~\cite{qvq-max} & - &\textbf{2.3}&2.9&6.1&6.0&5.6&7.8&94.9&2.4&2.5&---&75.8\\
Doubao-Seed-1.6~\cite{doubao-seed} & - &4.9&7.0&\textbf{3.5}&8.9&6.5&6.2&96.9&3.2&2.4&$\rm U$&80.3\\
GPT-4o-latest~\cite{gpt-4o} & - &5.7&5.1&7.1&7.7&6.3&\textbf{4.3}&98.1&2.9&1.4&---&79.9\\
Gemini 2.5 Pro$^\dagger$~\cite{gemini25} & - &7.0&\textbf{0.0}&7.5&8.1&\textbf{4.7}&4.9&\textbf{98.4}&\textbf{2.0}&\textbf{0.9}&---&\textbf{81.7}\\
Claude-Sonnet-4~\cite{Claude-Sonnet}& - &7.6&9.5&8.4&\textbf{5.3}&7.4&7.8&96.1&3.3&2.6&---&78.3\\
\noalign{\vskip 1pt}
\hline
\end{tabular}
}
\end{table*}

\subsection{Customized Context Construction}
\label{sec:3.3}
Our design enables customized contextual contents to simulate positional bias under various real-world scenarios: 
Multi-video understanding. The probe is inserted into a context with multiple videos.
Long video understanding. The probe is inserted at specific temporal positions within a natural long-form video.
Multi-modal interleaved input. The probe is inserted into an alternating sequence of text and video, a type of context commonly used in retrieval-augmented generation and multi-turn dialogues~\cite{multi-turn}. 
Template video background. Inspired by~\cite{cca}, 
a template video initialized with ImageNet mean pixel values is studied to isolate contextual effects.

\subsection{Positional Bias Metric}
\label{sec:3.4}
Considering the inherent capability variance in models, 
common used absolute accuracy is inadequate for fair cross-model positional bias comparison. To this end, 
we introduce the relative score (RS) metric, defined as:
\begin{equation}
	RS_{i} = \frac{S_{i}}{S_{meta}} \\
	\label{eq:eaem}
\end{equation}
where $S_{i}$ is the model accuracy when the probe is inserted at the $i$-th position, and $S_{meta}$ is the accuracy when the probe is provided standalone without any context. Based on this, positional bias is measured by combining both statistical metrics and morphological recognition. 

For statistical metrics, we propose: Position mean score, $P_{mean}$ = $mean(\{RS_{i}\}_{i=1}^{N})$, where $N$ is the number of evaluated positions, indicates the model's average performance across positions. Position range, $P_{ran}$ = $max(\{RS_{i}\}_{i=1}^{N})-min(\{RS_{i}\}_{i=1}^{N})$, measures the extent of the worst-case variation. Position variance, $P_{var}$ = $Var(\{RS_{i}\}_{i=1}^{N})$, evaluates the positional stability of the model.
To intuitively reflect the bias pattern, we use morphological recognition (\textit{MR}) to classify models into five types. Stable with milder bias (---). Neighbor preference ($\nearrow$). Head preference ($\searrow$). Lost in the middle ($\rm U$). Volatile score ($\rm W$). Classification is based on polynomial fitting, detailed in \cref{sec:MR}.
\section{Experiments}
\label{sec:exp}

We first detail the evaluation protocol, and then present evaluations of a wide range of LVLMs. Next, we analyse the effects of context length, context type, and model size.

\subsection{Evaluation Protocol}
We conduct a comprehensive investigation of 27 LVLMs using Video-LevelGauge, including 6 commercial models, i.e., Gemini 2.5 Pro~\cite{gemini25} and QVQ-Max~\cite{qvq-max}; 21 open-source LVLMs covering unified models like InternVL3~\cite{internvl3}, long video models like Video-XL2~\cite{video-xl2}, specific optimized models like VideoRefer~\cite{videorefer}, multi-modal reasoning models like GLM-4.5V~\cite{glm45v}, and two-stage methods like LLoVi~\cite{llovi}.
Specifically, we evaluate ten uniformly distributed positions across the video context.
For each probe, the context is built using nine same type collected videos, resulting in synthetic contexts with an average duration of 7.2 minutes, which are sampled following official implementations of each model. 
Unless specified, we sample 6 frames per probe and test the model's response by inserting them into different positions, ensuring consistent input of probe frames to the model to isolate sampling impact.
Details are provided in \cref{sec:appendix}.

\subsection{Evaluation of SOTA LVLMs}
As shown in \cref{tab:main}, commercial models generally exhibit milder positional bias compared to open-source models. In particular, Gemini 2.5 Pro~\cite{gemini25} demonstrates minimal positional bias (with a small $P_{ran}$ and $P_{var}$, and a stable \textit{MR}), followed by GPT-4o~\cite{gpt-4o}, Doubao-Seed-1.6~\cite{doubao-seed}, and QVQ-Max~\cite{qvq-max}, commercial models known for their long-context capabilities.
Notably, Qwen-VL-Max~\cite{qwenvl}, as an image model, shows more pronounced positional bias. We speculate that this difference may be attributed to the longer training contexts and larger parameters of commercial models, which can mitigate bias as analysed in \cref{sec:4.3}.

Among the open-source models, two-stage methods that utilize LLMs to interpret sequential frame descriptions exhibit a U-shaped performance, as found in previous work~\cite{lostmiddle}. 
The state-of-the-art reasoning models~\cite{glm45v,mimovl} exhibit minimal positional bias, particularly GLM-4.5V~\cite{glm45v} with a parameter scale of 10.8 billion (108B). Meanwhile, it can be observed that reasoning patterns can alleviate the positional bias issue to a certain extent (\cref{fig:thinking}). 
Besides, InternVL3~\cite{internvl3}, Video-LLaMA3~\cite{videollama3}, and LLaVA-Video~\cite{llava-video} are also impressive in terms of reduced positional bias. Video-XL2~\cite{video-xl2}, a model optimized for long videos, also exhibits relatively small positional bias. However, LongVA~\cite{longva} and LongVILA~\cite{longvila} fail to achieve consistently robust performance across the entire sequence. This suggests that the effectiveness of long context training needs to be verified, as analysed in \cref{sec:4.3}. 

We also observe that models tend to exhibit lower positional bias on tasks in which they excel. For example, Qwen2.5-VL~\cite{qwen25vl} shows reduced positional bias on the OCR task compared to its bias on other tasks, while MiniGPT4-Video~\cite{minigpt4videos} displays more noticeable bias in Optical Character Recognition (OCR) and Attribute Perception (AP) tasks. Unexpectedly, T-Star~\cite{T-star}, as a context search algorithm, exhibits positional fluctuations. We attribute this to the relatively short duration of our probes. T-Star may fail to include probe frames in its first sampling round, leading to sub-optimal context search in subsequent steps.

Overall, there remains considerable room for improvement in addressing positional bias, and we discuss potential directions in \cref{sec:disc}.
\begin{figure*}[t]
    \centering
    \begin{subfigure}[b]{0.33\textwidth}
        \centering
        \includegraphics[width=\textwidth]{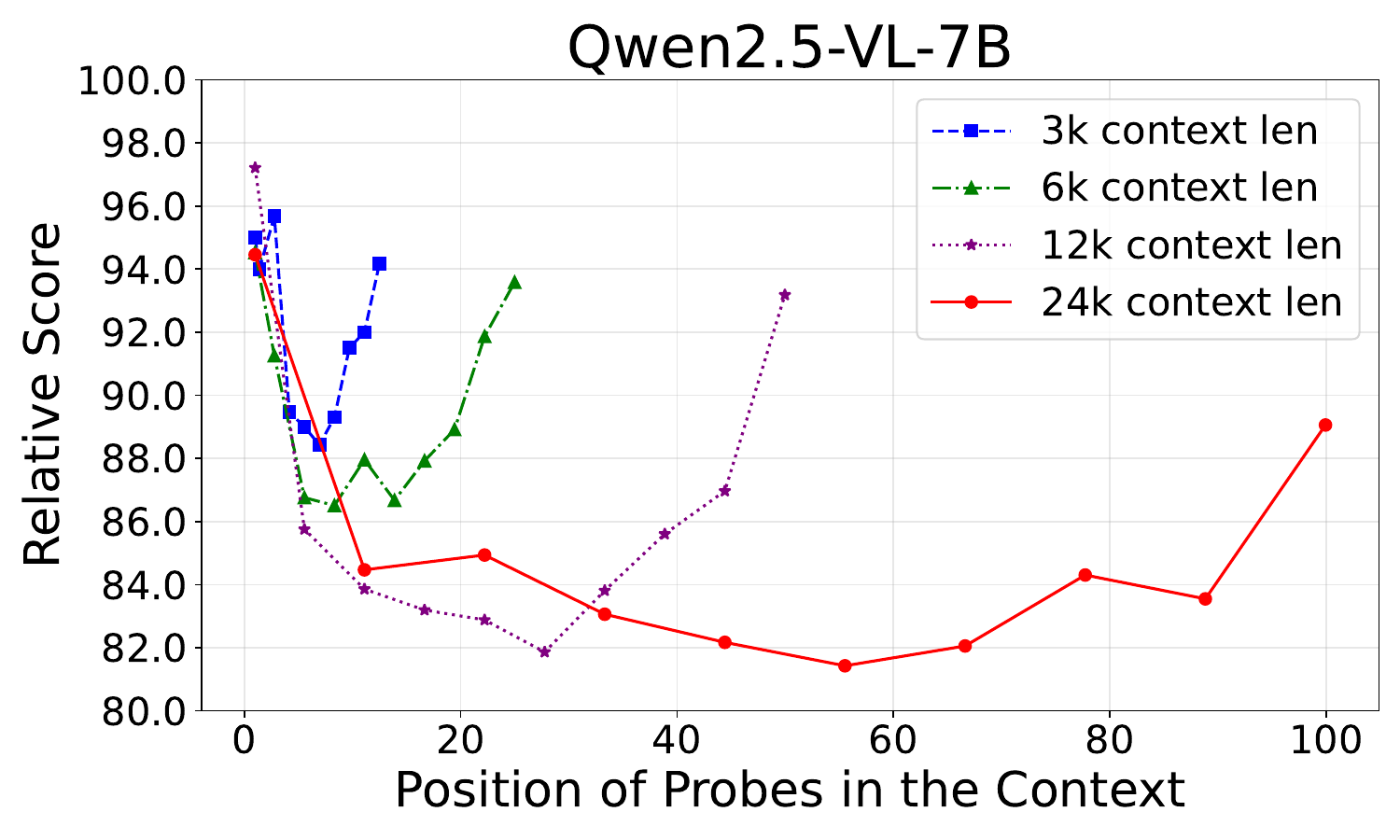} 
    \end{subfigure}
    \hfill
    \begin{subfigure}[b]{0.33\textwidth}
        \centering
        \includegraphics[width=\textwidth]{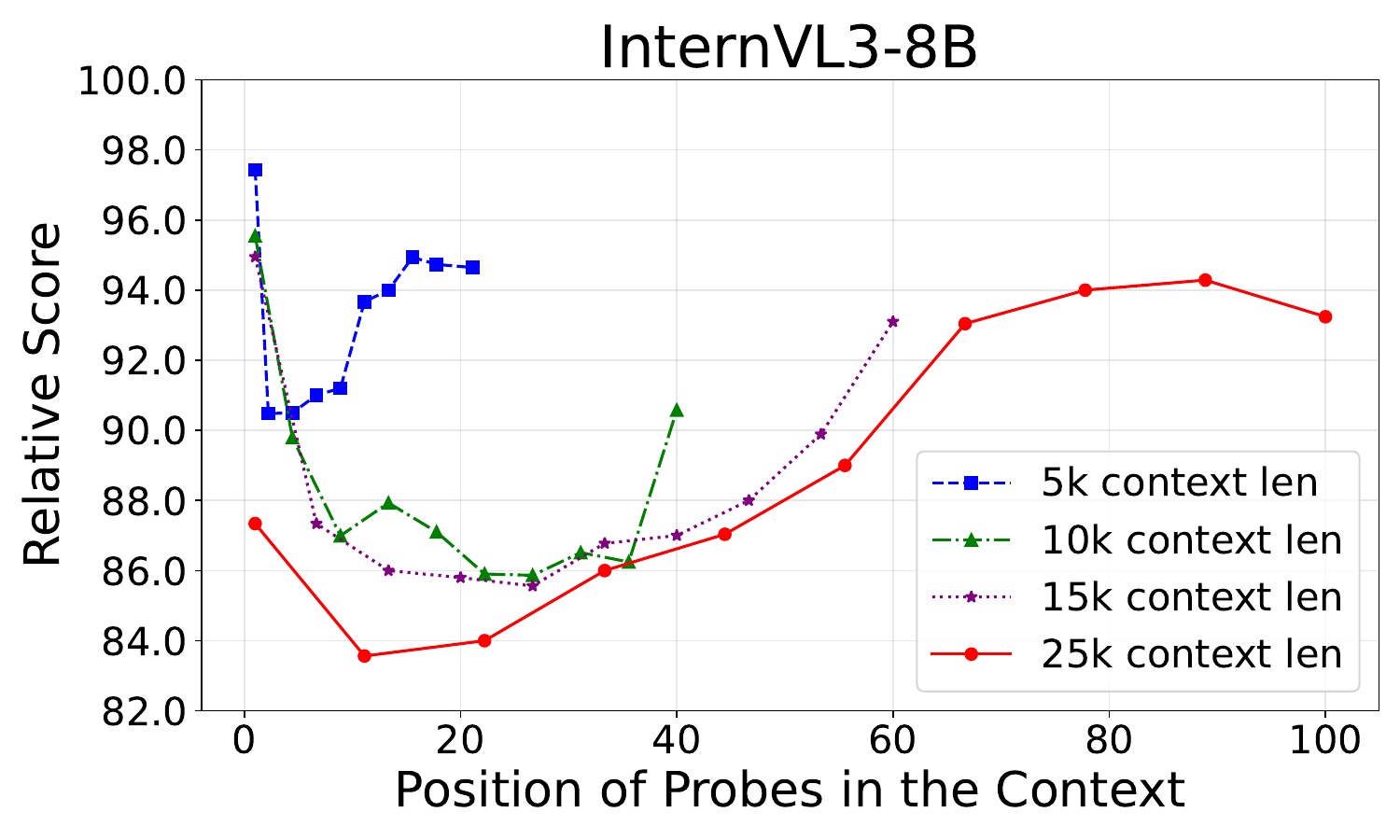}
    \end{subfigure}
    \hfill
    \begin{subfigure}[b]{0.33\textwidth}
        \centering
        \includegraphics[width=\textwidth]{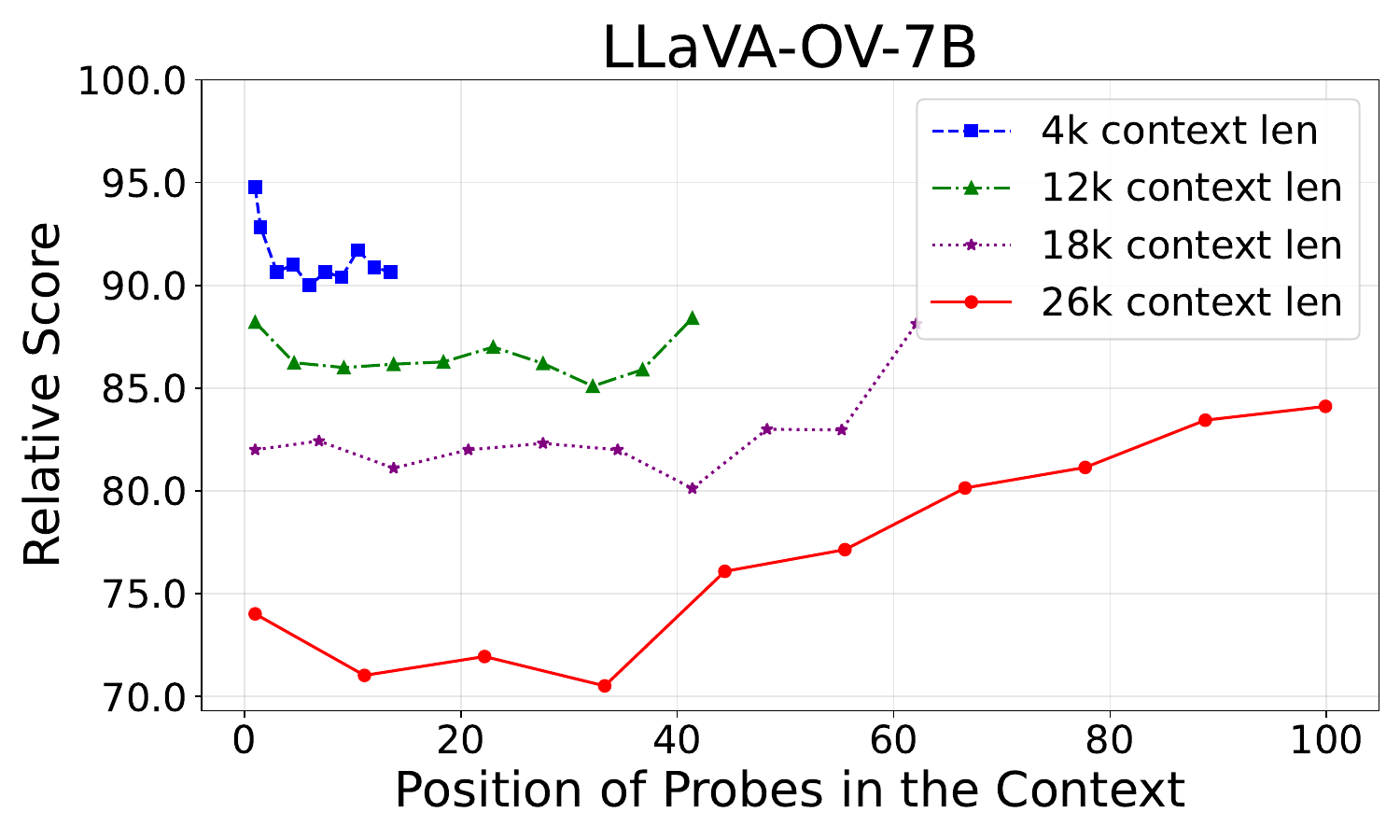}
    \end{subfigure}
    
    \begin{subfigure}[b]{0.33\textwidth}
        \centering
        \includegraphics[width=\textwidth]{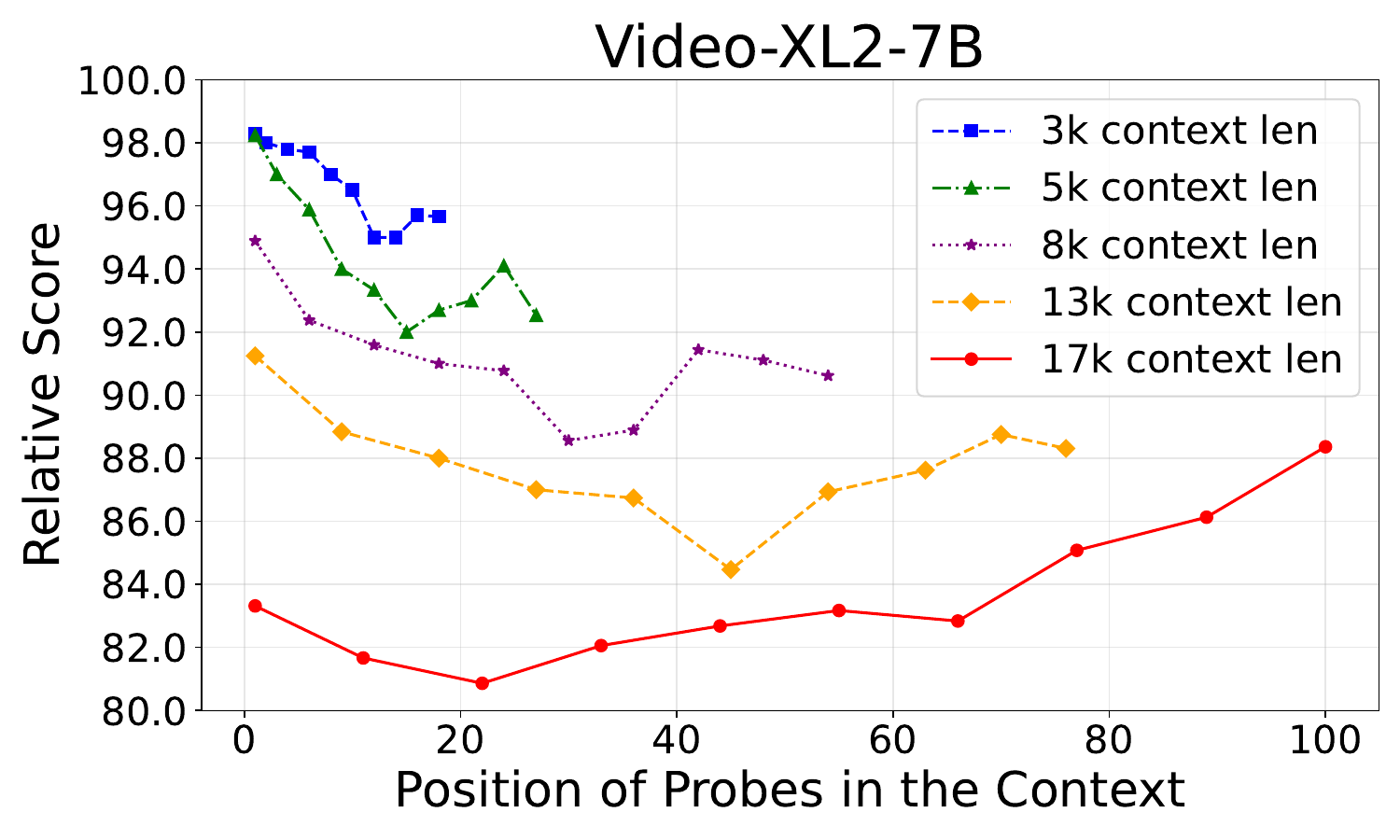}
    \end{subfigure}
    \hfill
    \begin{subfigure}[b]{0.33\textwidth}
        \centering
        \includegraphics[width=\textwidth]{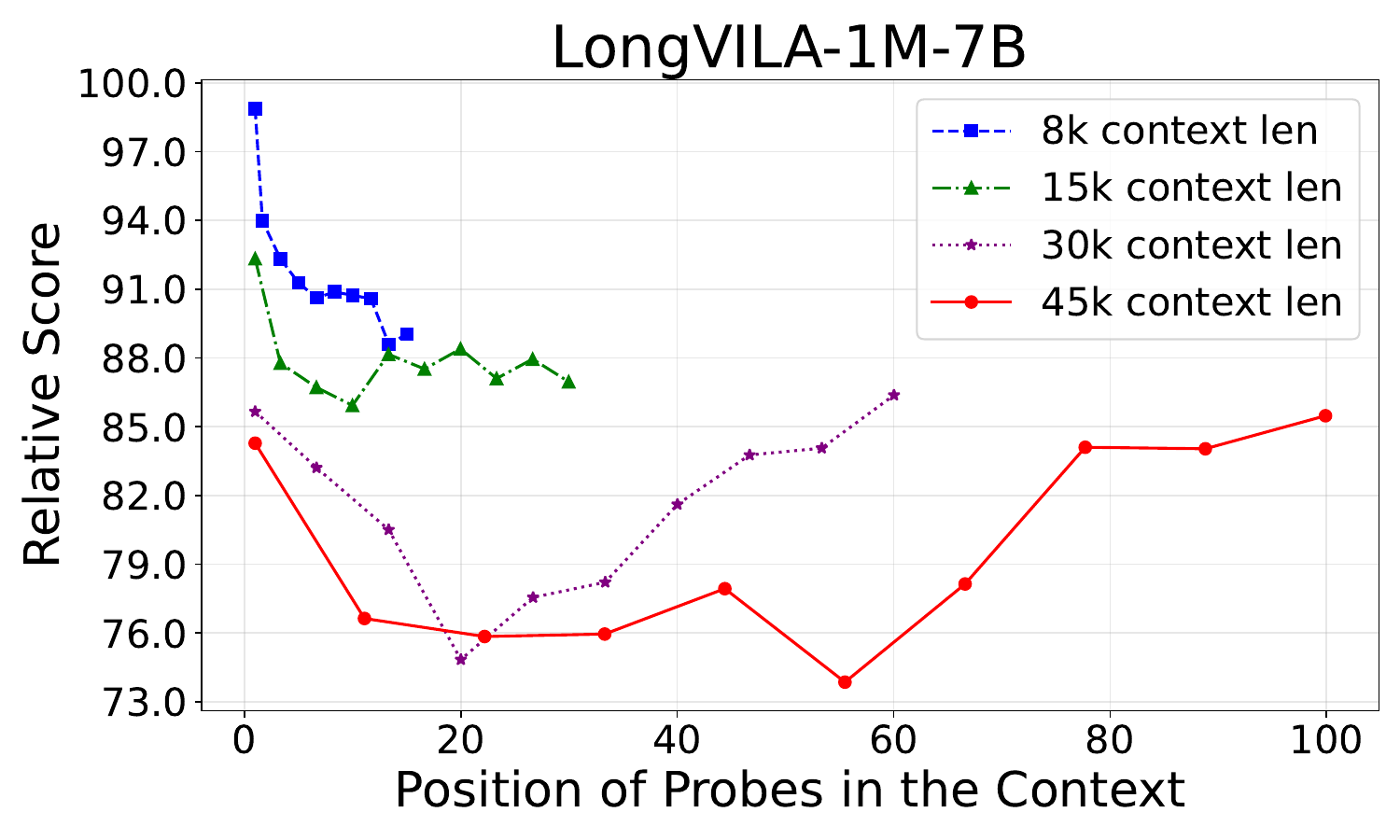}
    \end{subfigure}
    \hfill
    \begin{subfigure}[b]{0.33\textwidth}
        \centering
        \includegraphics[width=\textwidth]{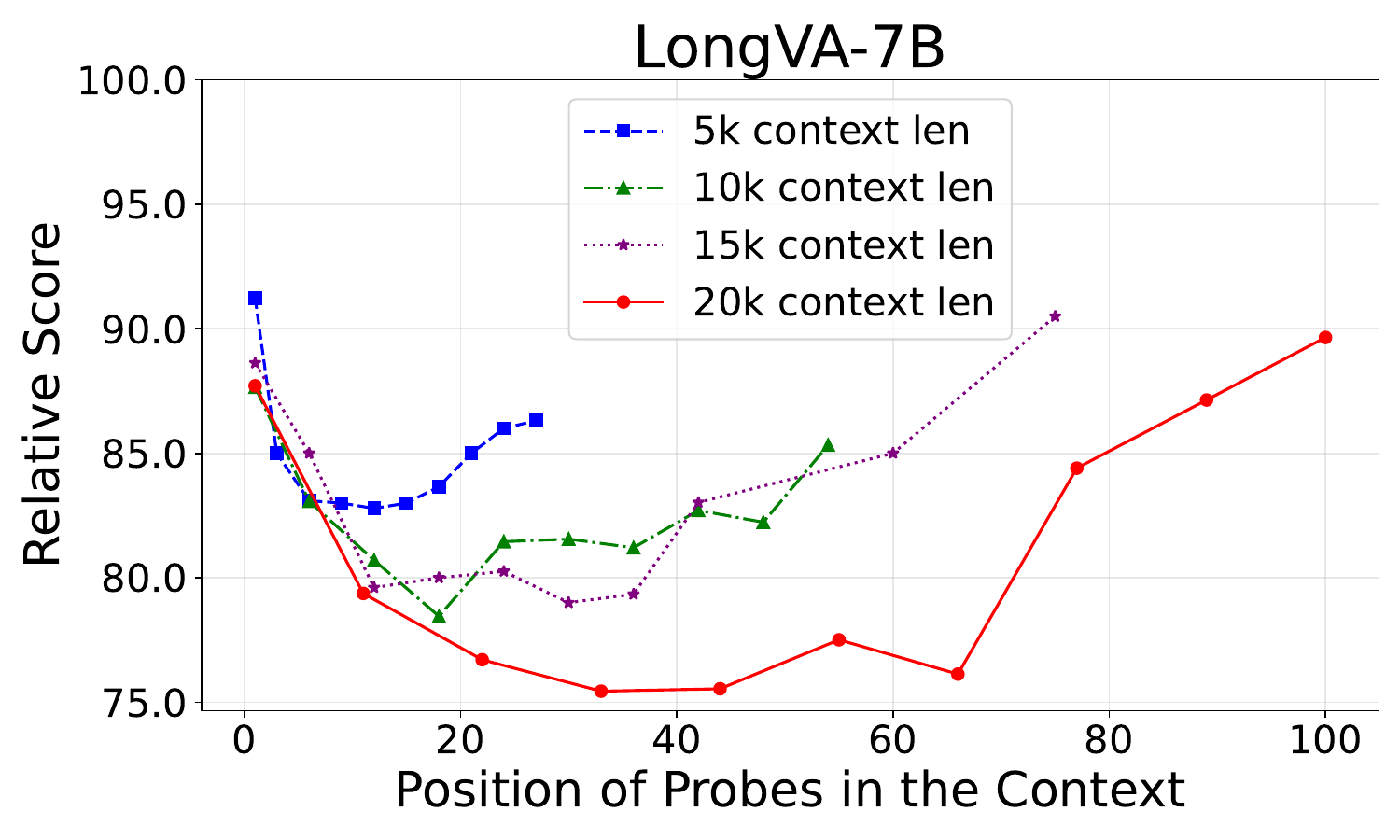}
    \end{subfigure}
    
    \begin{subfigure}[b]{0.33\textwidth}
        \centering
        \includegraphics[width=\textwidth]{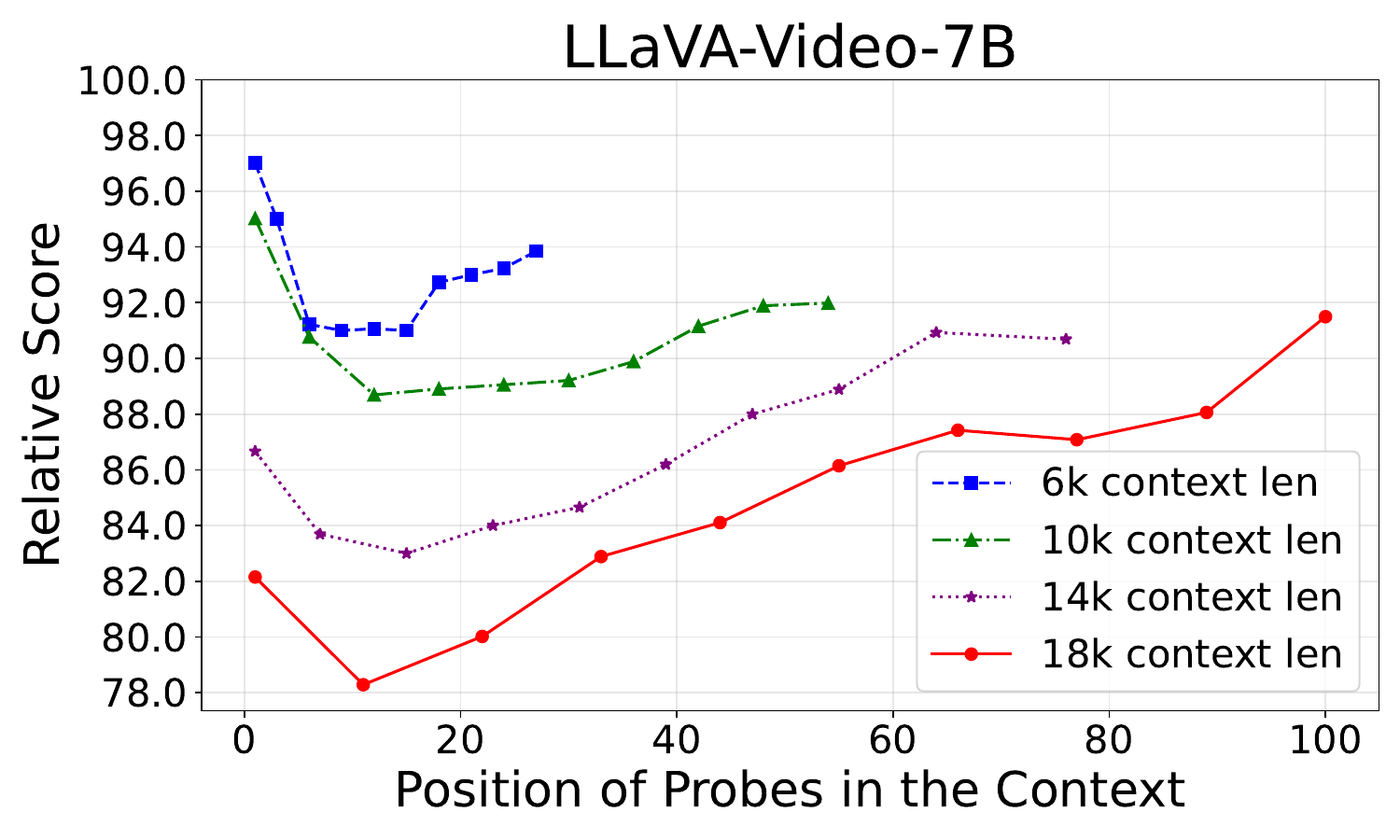}
    \end{subfigure}
    \hfill
    \begin{subfigure}[b]{0.33\textwidth}
        \centering
        \includegraphics[width=\textwidth]{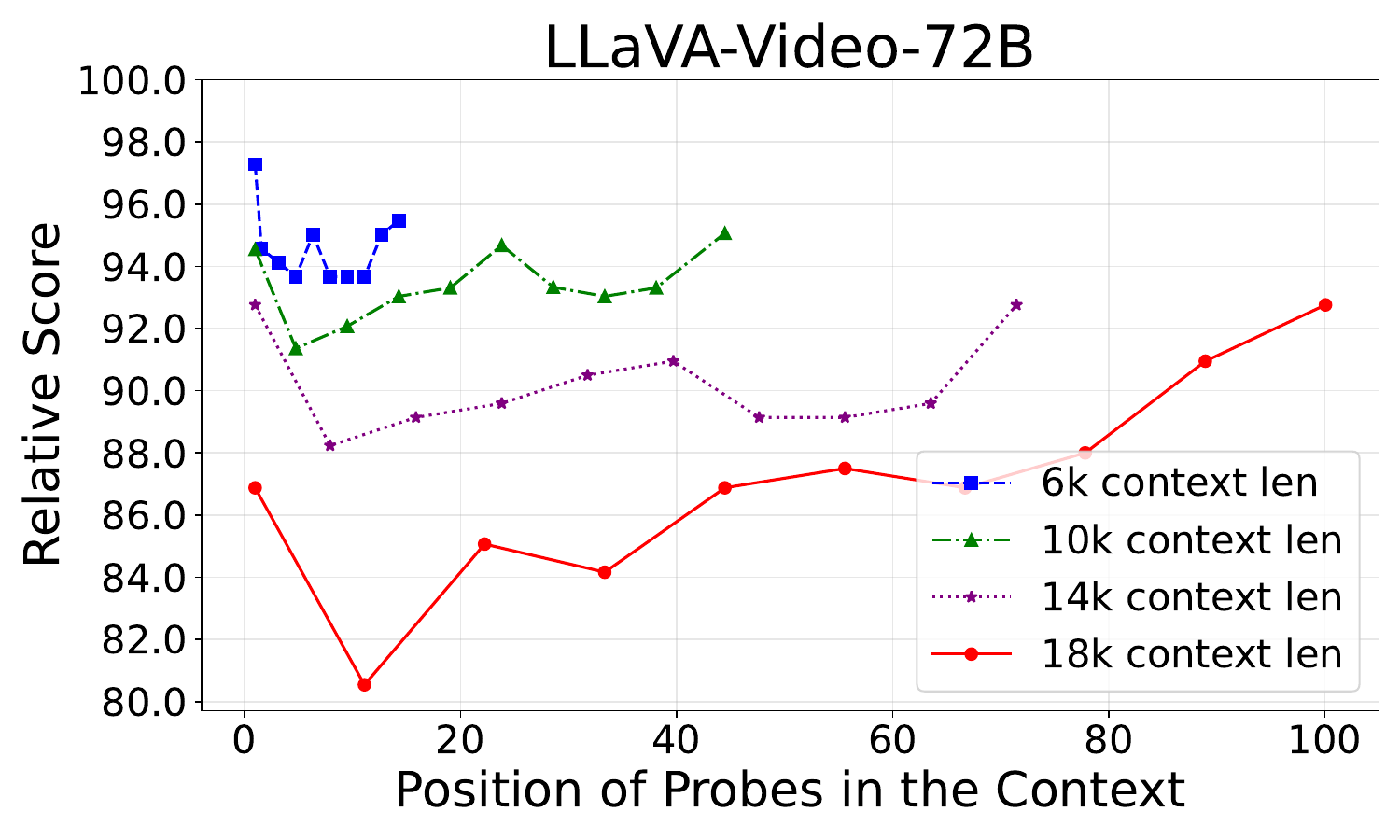}
    \end{subfigure}
    \hfill
    \begin{subfigure}[b]{0.33\textwidth}
        \centering
        \includegraphics[width=\textwidth]{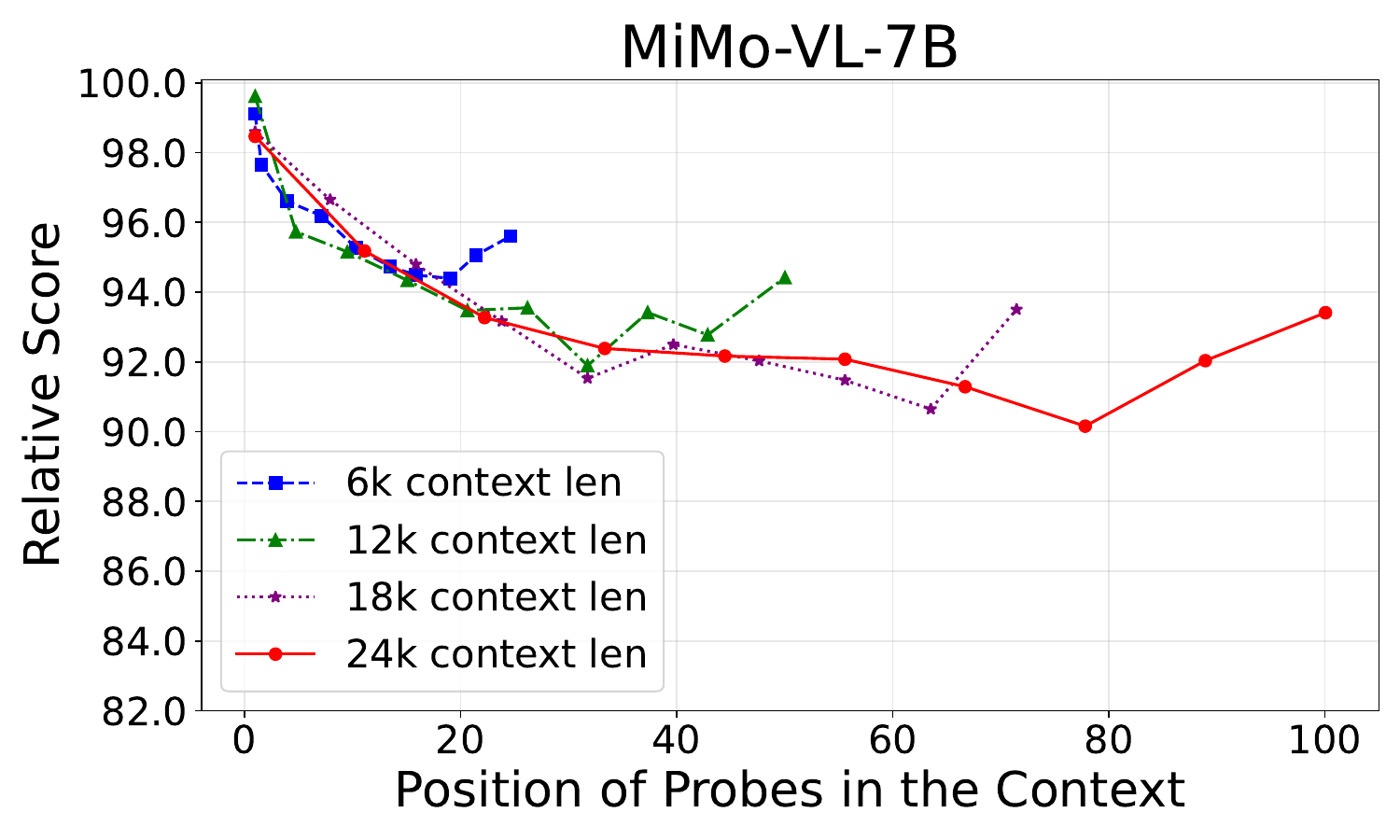}
    \end{subfigure}
    \caption{
    \textbf{Effect of context length on positional bias.} Each plot illustrates how positional bias manifests across different context lengths on a specific LVLM. The horizontal axis denotes the position of probes in the context, and the vertical axis shows the model’s relative score. Colored curves represent different context lengths. Positional bias is prevalent across various context lengths. It tends to intensify as the context length increases, and the patterns of bias may shift with increasing context.
    }
    \label{fig:context_len}
\end{figure*}

\subsection{Further Analysis}
\label{sec:4.3}
We further investigate the effect of context length, context variation, and model scale, providing actionable insights. % for future model enhancement.
\subsubsection{Effect of Context Length.}
Models tend to struggle with long-context scenarios, possibly due to inference lengths exceeding the training horizon~\cite{scalinglaw}. As shown in \cref{fig:context_len}, we investigate the effect of context length on positional bias across eight representative models, including models built upon long-context LLMs, such as InternVL3 and LongVA, and models specifically optimized via long video training, such as LongVILA-1M.

\textbf{Finding 1: Positional bias is prevalent across various context lengths and tends to intensify as the context length increases, accompanied by shifts in bias patterns.} 
We observe that models such as Video-XL2~\cite{video-xl2}, LongVILA~\cite{longvila}, and LLaVA-OV~\cite{llavaonevision} undergo transitions among three patterns of positional bias: a head preference ($\searrow$) in shorter contexts, followed by a phase of lost in the middle ($\rm U$), and eventually a shift toward neighbor bias ($\nearrow$) in longer contexts. Moreover, models like Qwen2.5-VL~\cite{qwen25vl} and LongVA~\cite{longva} exhibit U-shaped performance across different lengths. This suggests that LVLMs may inherit the biases of the language models. Optimization of positional encoding~\cite{qwen25vl} and the effectiveness of long video post-training~\cite{video-xl2} should be verified through positional bias evaluation. Their overall performance gains on long videos may stem less from improved comprehension of the entire sequence, and more from an enhancement of visual understanding due to exposure to more video data.
It is recommended to pay closer attention to positional bias, particularly when developing long video models.
\begin{figure*}[t]
    \centering
    \includegraphics[width=0.21\linewidth]{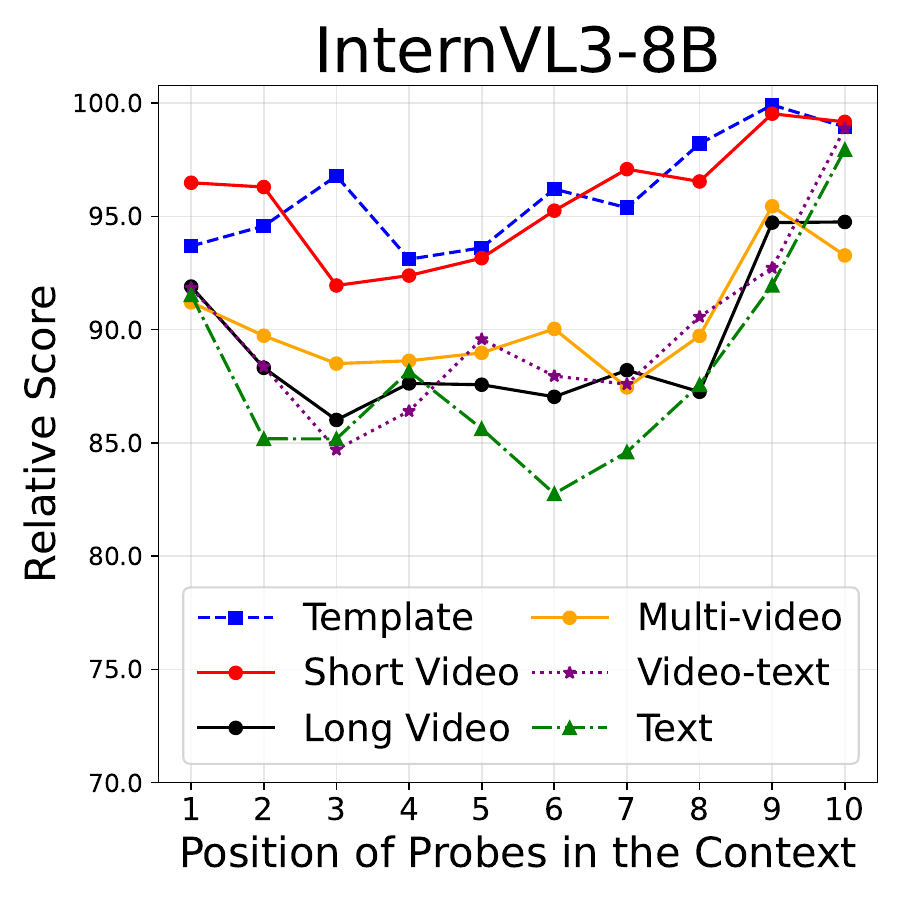}
    \hfill
    \includegraphics[width=0.11\linewidth]{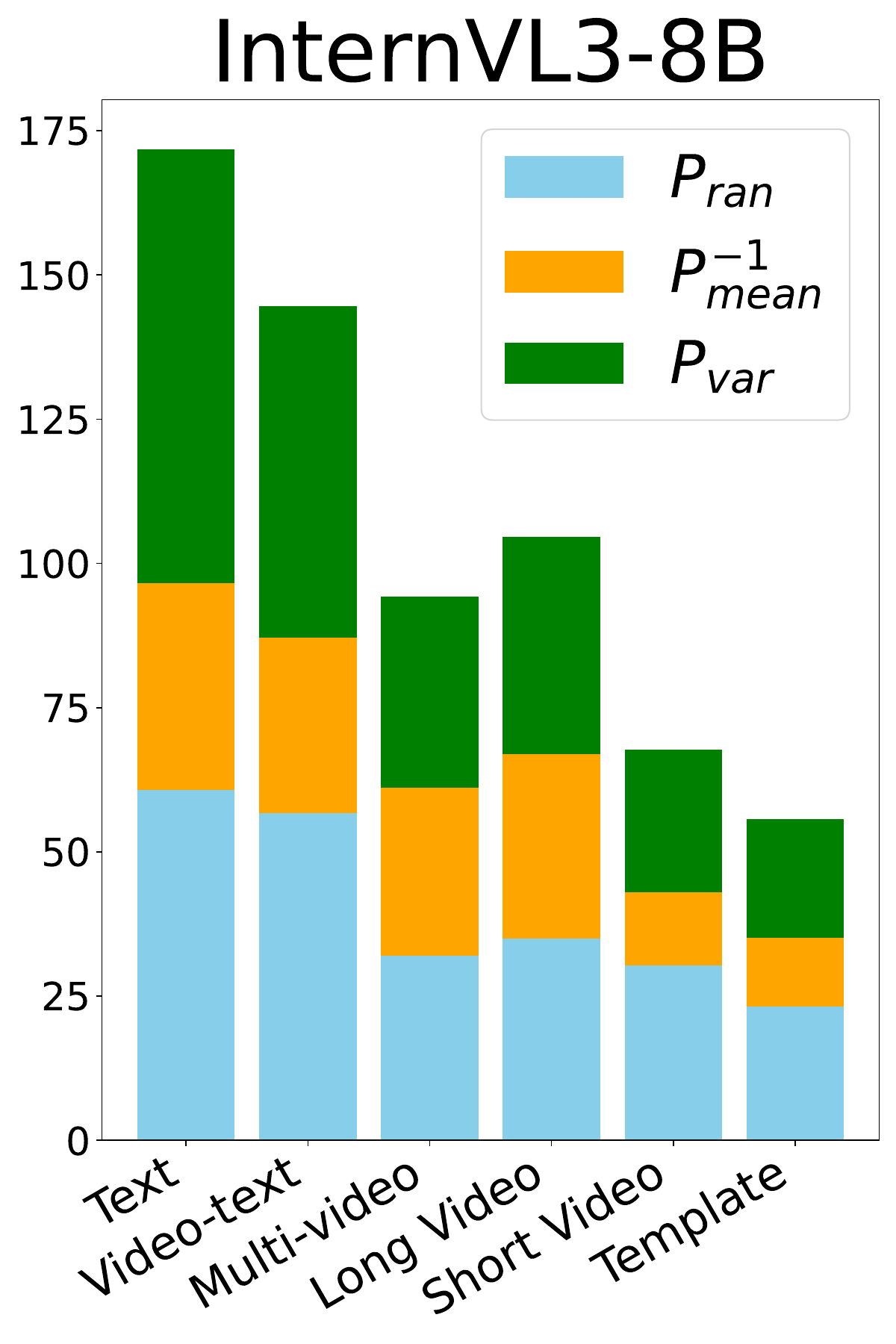}
    \hfill
    \includegraphics[width=0.21\linewidth]{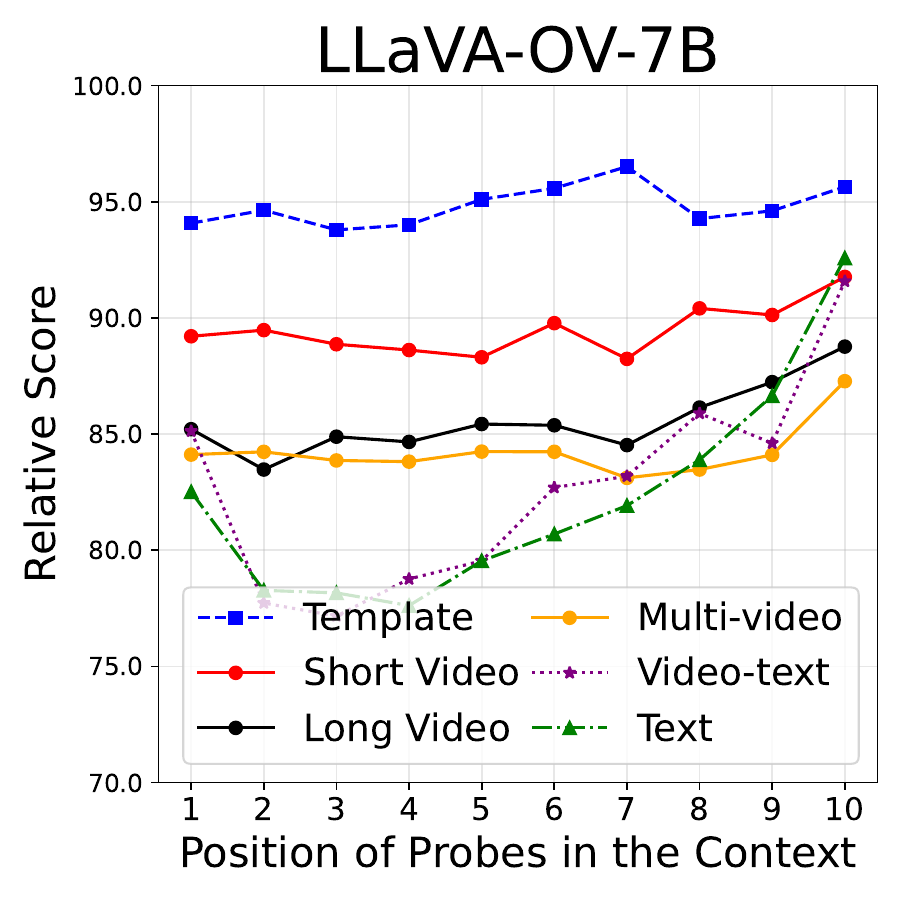}
    \hfill
    \includegraphics[width=0.11\linewidth]{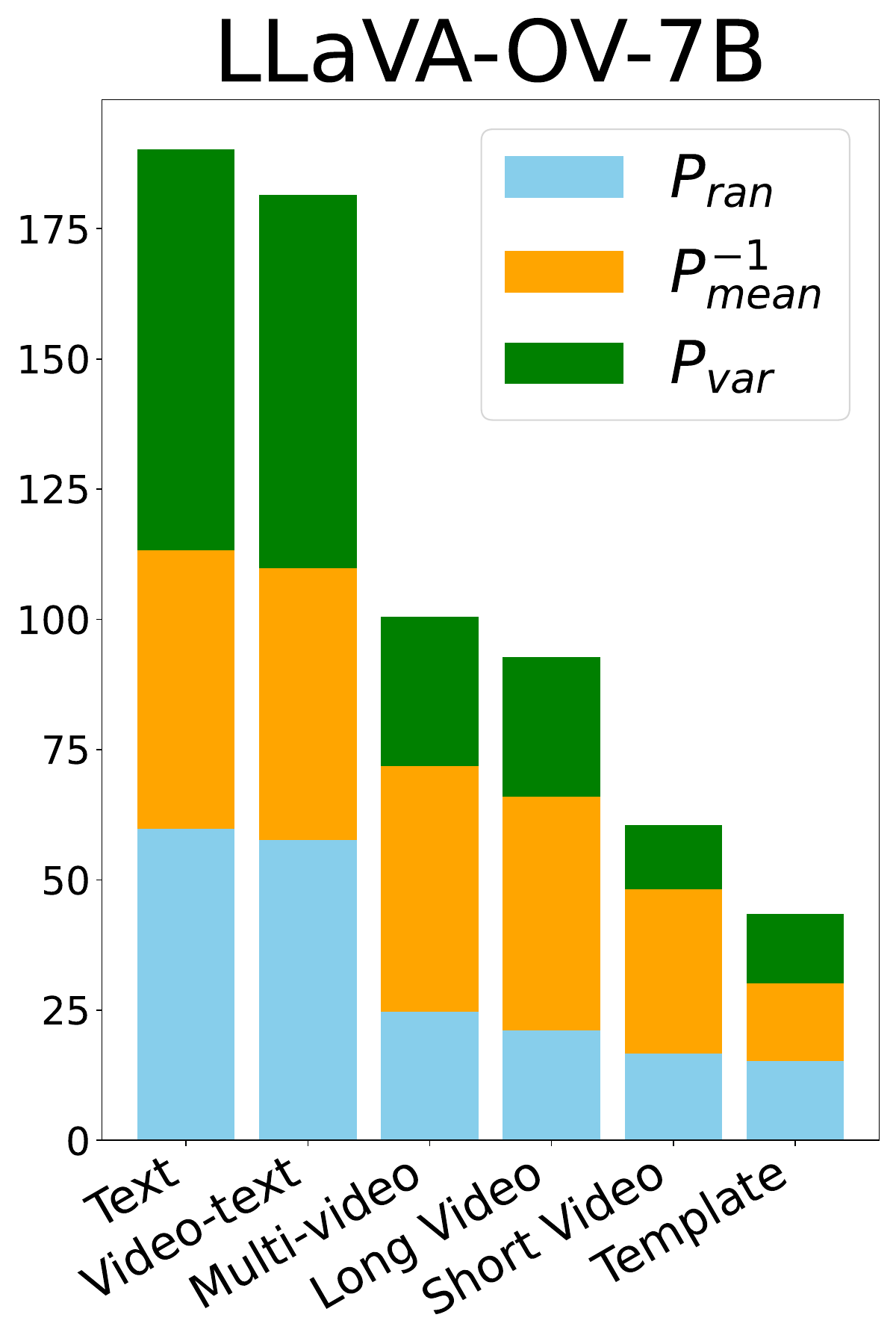}
    \hfill
    \includegraphics[width=0.21\linewidth]{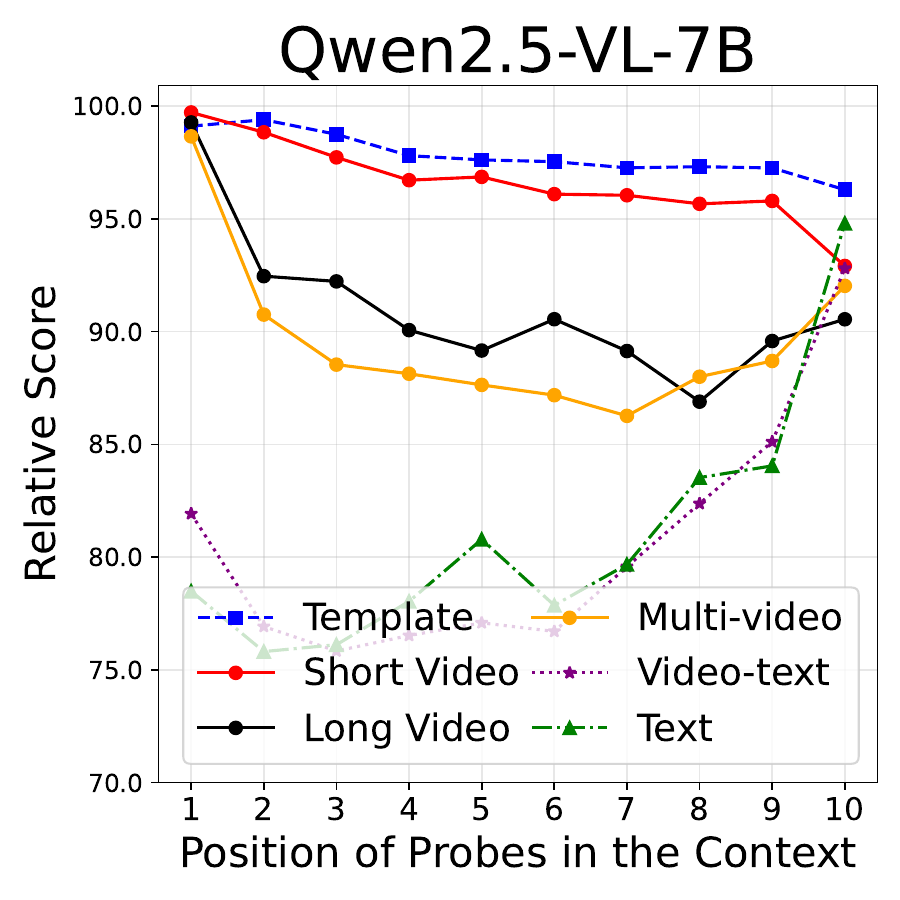}
    \hfill
    \includegraphics[width=0.11\linewidth]{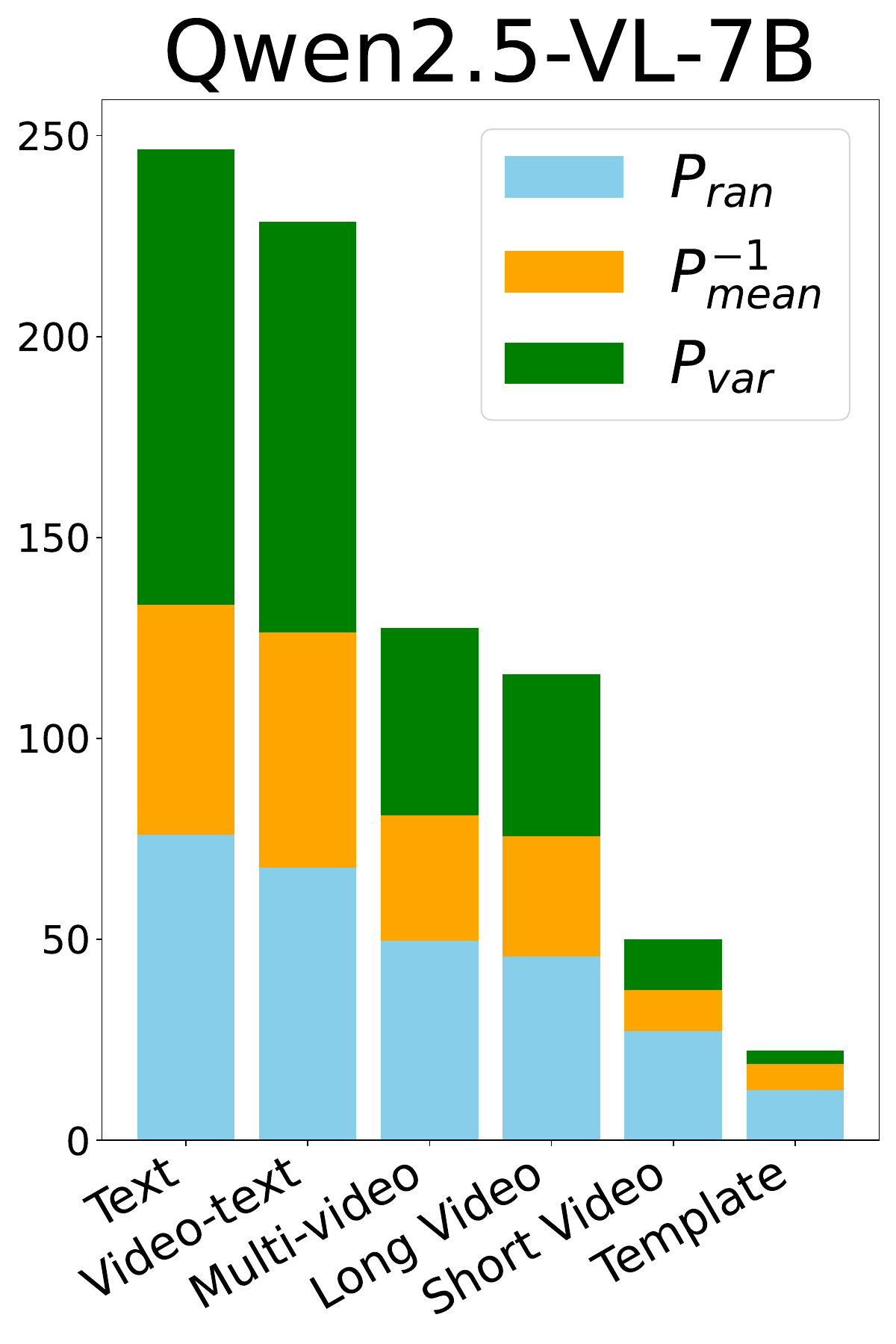}
    \caption{\textbf{Effect of context type on positional bias.} 
    For each model, we visualize the variation in relative scores across probe positions under six types of context (left), and quantify the severity of positional bias using the composite value of three proposed metrics (right), where longer bars indicate more pronounced bias. LVLMs exhibit milder positional bias in simple contexts, such as template videos, and degrade significantly in complex contexts, such as interleaved video-text inputs.
    }
    \label{fig:context_type}
\end{figure*}

\subsubsection{Various Customized Contexts.}
Our benchmark supports flexible customized contexts. To explore positional bias under different real-world scenarios, we experiment with six distinct context types while keeping the context length fixed. These include: (1) template context, (2) short video context, (3) long video context, (4) multi-video context, (5) long-text context, and (6) interleaved video-text context. These contexts are sourced from public datasets~\cite{obelicsopen,lvbench}, detailed in \cref{sec:context}.

\textbf{Finding 2: LVLMs exhibit more pronounced positional bias in complex context scenarios.}
As shown in \cref{fig:context_type}, models generally maintain high performance and exhibit milder positional bias in simpler contexts, such as template and short video context settings. However, as the contextual complexity increases, such as in long video and multi-video scenarios, positional bias becomes more pronounced. Notably, the most severe bias emerges when textual contents are introduced into the context, such as in interleaved video-text and long-text contexts. We attribute this to the lack of training on long mixed-modal context data. 
These results underscore the vulnerability of LVLMs to positional bias when exposed to mixed-modal inputs, which are prevalent in multi-turn dialogue or retrieval-augmented generation scenarios.

\subsubsection{Effect of Model Size.}
\begin{figure}[t] 
    \centering  
    \begin{subfigure}[b]{0.23\textwidth}
        \centering
        \includegraphics[width=\textwidth]{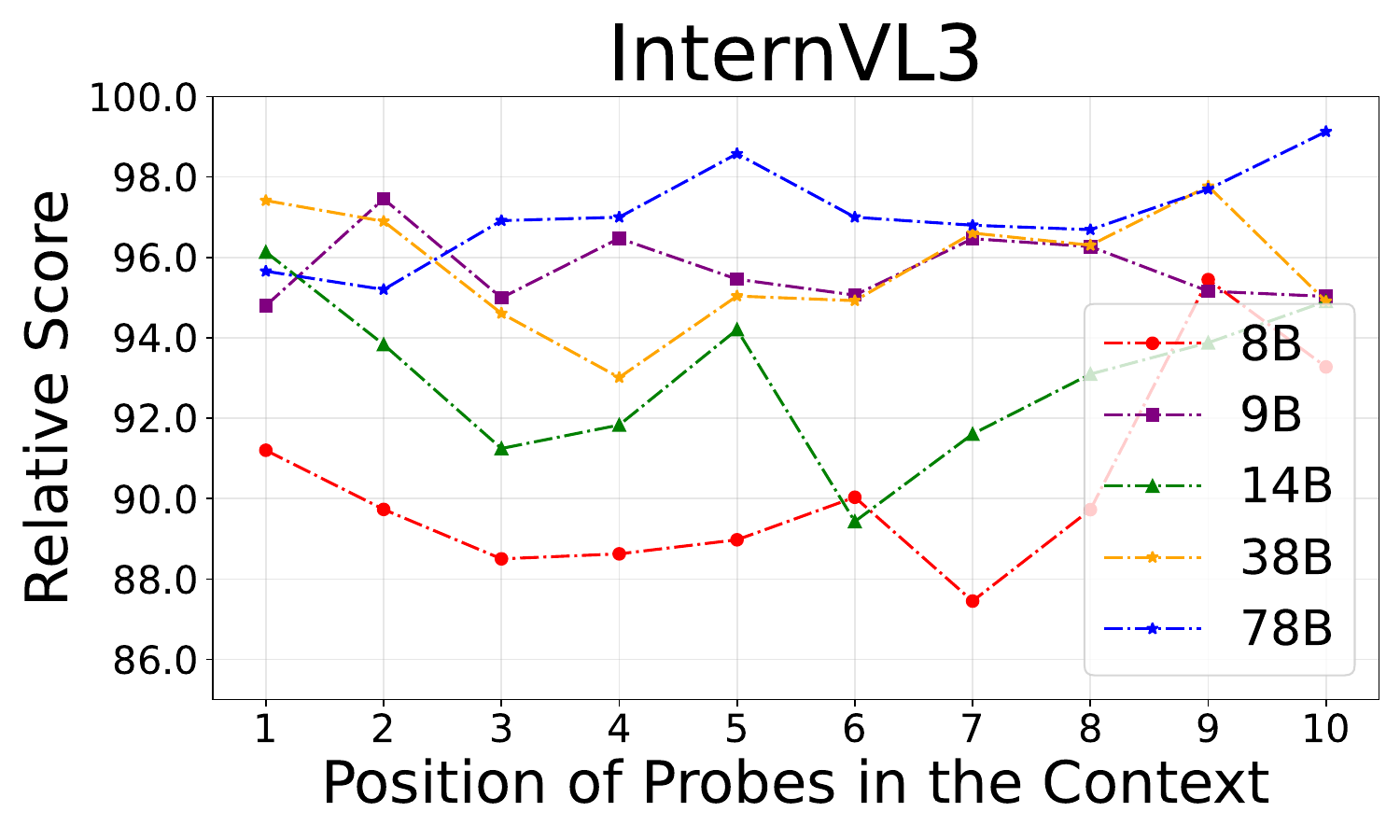} 
    \end{subfigure}
    \begin{subfigure}[b]{0.23\textwidth}
        \centering
        \includegraphics[width=\textwidth]{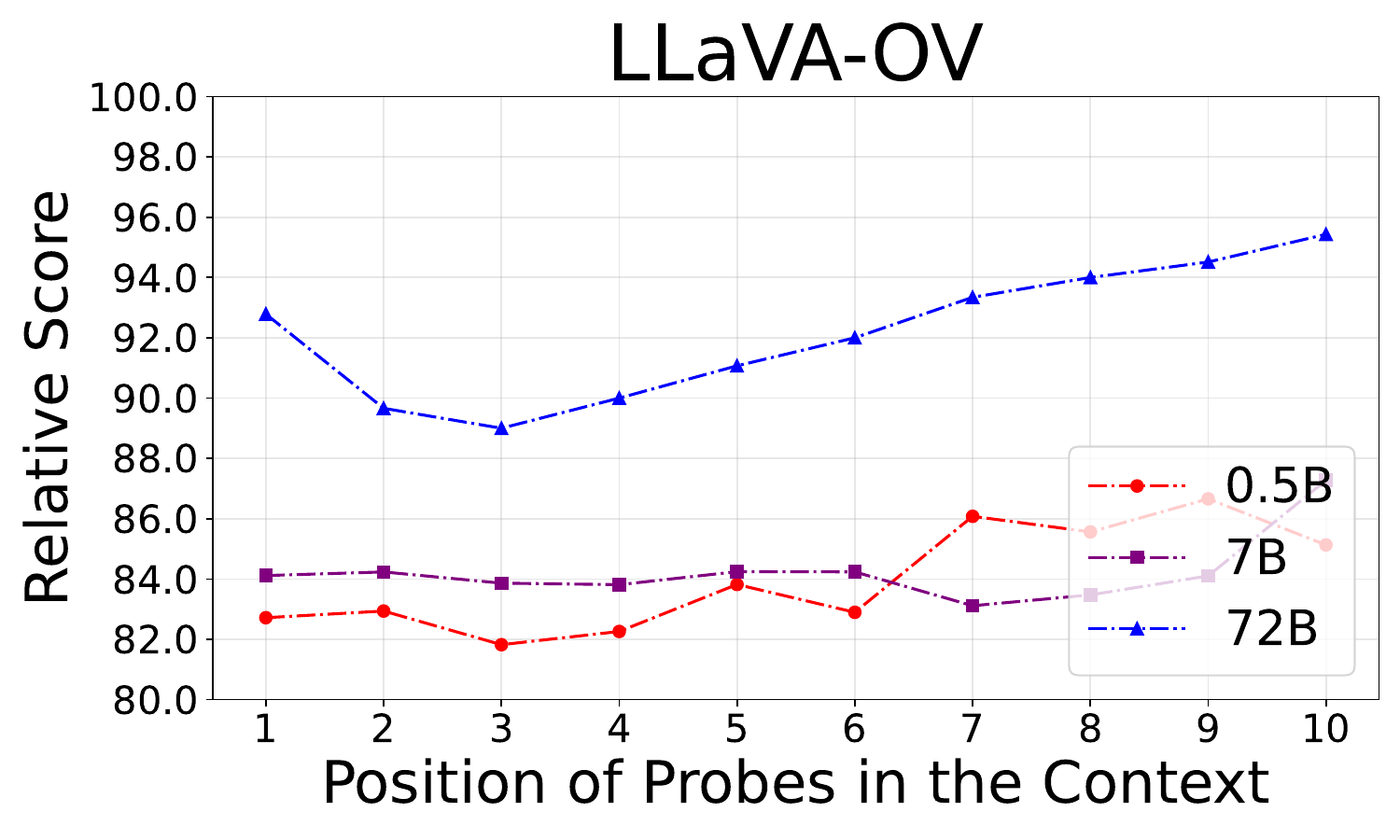} 
    \end{subfigure}
    
    \begin{subfigure}[b]{0.23\textwidth}
        \centering
        \includegraphics[width=\textwidth]{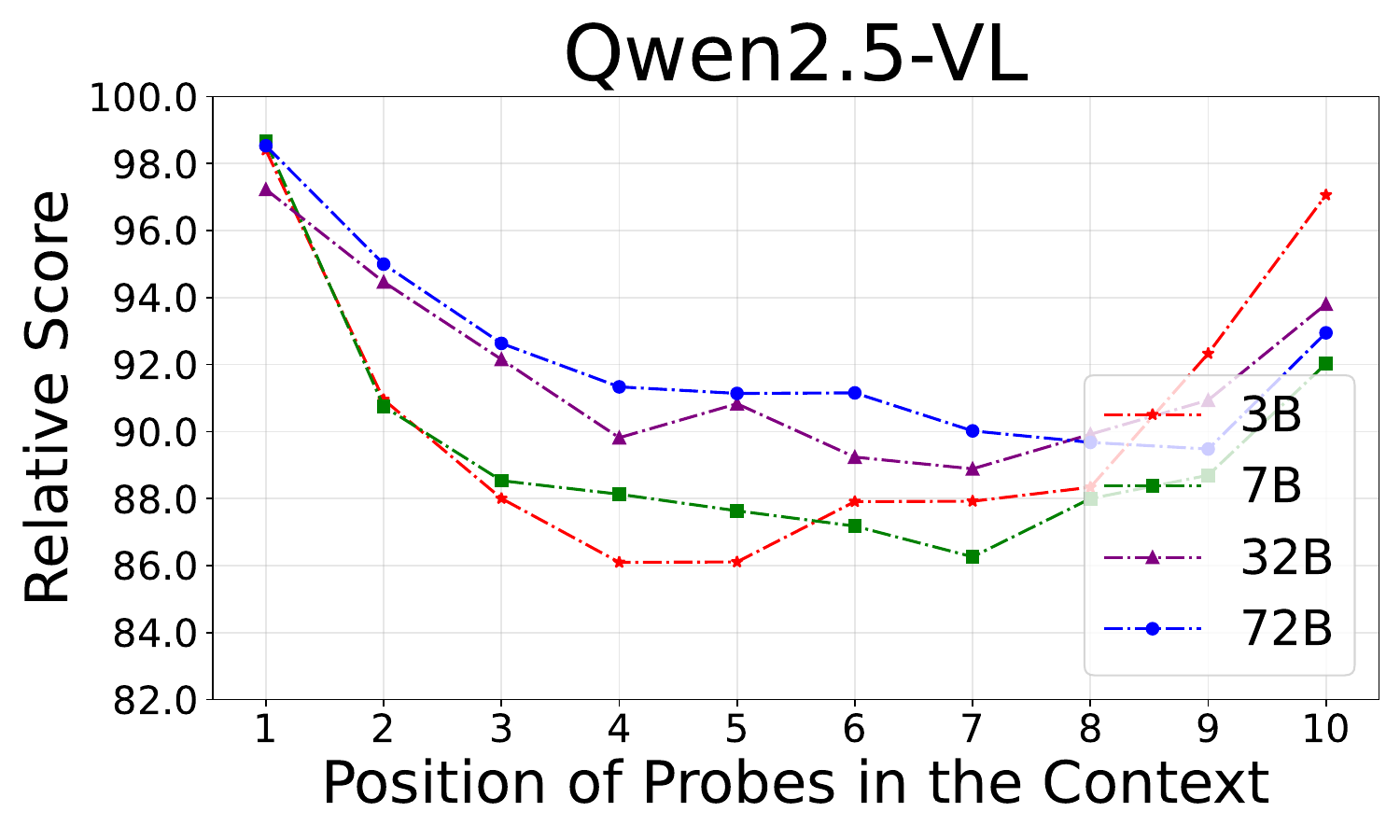}
    \end{subfigure}
    \begin{subfigure}[b]{0.23\textwidth}
        \centering
        \includegraphics[width=\textwidth]{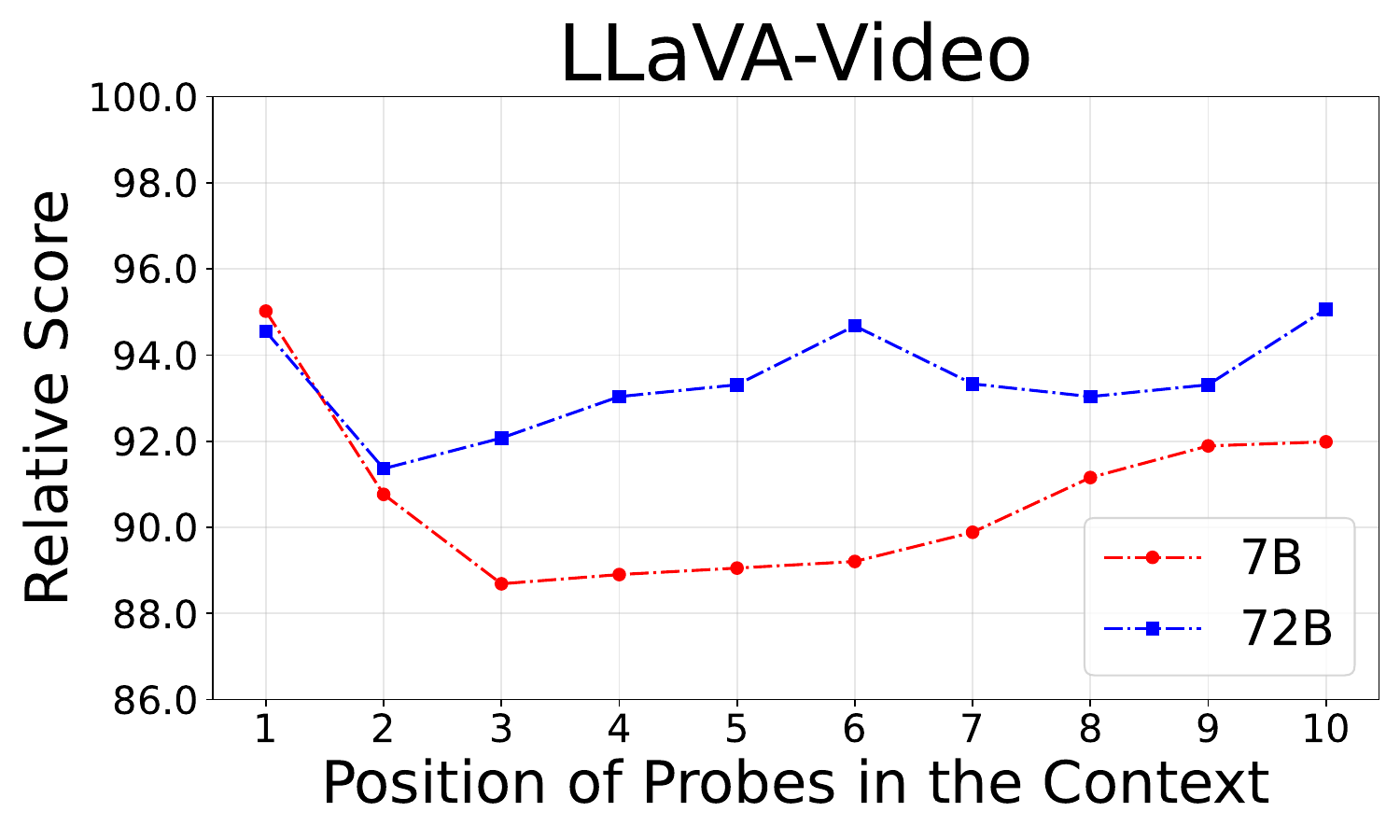} 
    \end{subfigure}
       
    \caption{\textbf{Effect of model size on positional bias.} Larger variants exhibit more stable and superior performance across the entire sequence than their smaller counterparts.} 
    \label{fig:model_size} 
\end{figure}

According to the scaling law~\cite{scalinglaws}, increasing the number of model parameters generally improves performance across a wide range of capabilities. To investigate the effect of model size on positional bias, we conduct experiments on four representative models, as presented in \cref{fig:model_size}.

\textbf{Finding 3: Positional bias is significantly alleviated as model size increases, consistent with scaling law observed in other capabilities.} 
Within the same model series, larger variants exhibit more stable performance across the entire sequence, as visualized by flatter bias curves and quantified by lower $P_{var}$ in \cref{tab:main}. They also show greater resilience to long-context interference, reflected by consistently higher relative scores. Our experiments show that the superior overall performance of larger models on long video tasks stems not only from enhanced comprehension capabilities, but also from their improved handling of blind positions where small ones are more prone to positional bias.

\subsubsection{Open-ended Questions.}
Video-LevelGauge contains 120 open-ended descriptive questions, where LVLMs are instructed to describe a mentioned scene. We use GPT-4~\cite{gpt4} to evaluate model responses by comparing them against annotations. 
As shown in \cref{tab:caption-exp}, we observe slightly more pronounced positional bias when using descriptive tasks compared to MCQA formatted evaluation. We attribute this to descriptive tasks requiring more fine-grained and comprehensive perceptions, whereas multiple-choice questions typically target the perception of a specific object or action, making them relatively simple. As a result, portions of potential positional bias may go undetected in MCQA settings. 

\begin{table}[t]
\centering
\caption{\textbf{Evaluation on open-ended descriptive questions.}}
\resizebox{1.0\columnwidth}{!}{
\begin{tabular}{lccccc}
\hline
\noalign{\vskip 2pt}
Models & Model size & $P_{mean}$ & $P_{ran}$ & $P_{var}$ &$S_{meta}$ \\
\hline
\noalign{\vskip 2pt}
InternVL3~\cite{internvl3} & 8B &90.0&8.2&7.1&70.7\\
InternVL3~\cite{internvl3} & 38B &94.8&4.8&5.3&74.8\\
LLaVA-Video~\cite{llava-video} & 7B &90.3&7.9&4.9&70.0\\
LLaVA-OV~\cite{llavaonevision} & 7B &91.7&4.6&5.1&66.8\\
Qwen2.5-VL~\cite{qwen25vl} & 7B &87.5&9.4&13.8&68.3\\
Qwen2.5-VL~\cite{qwen25vl} & 32B &92.5&9.3&7.2&72.8\\
Video-XL2~\cite{video-xl2} & 7B &90.6&7.4&5.4&71.4\\
\hline
\end{tabular}
}
\label{tab:caption-exp}
\end{table}

\subsubsection{Discussion.}
\label{sec:disc}
Contextual positional bias, an inconspicuous but critical issue of LVLMs, may stem from biases in the training data, as data can exhibit serial position effects due to humans' first-and-last preference. Our analysis suggests that beyond scaling up model parameters, several strategies hold promise for mitigating positional bias, including long video training and enhanced context search algorithms. In addition, we observe amplified positional bias in multi-modal input scenarios, highlighting the potential of training with interleaved video-text data and developing cross-modal context search algorithms. 
It tends to worsen with longer context lengths; thus, excellent video token compression may not only improve efficiency but also mitigate bias issues.
Finally, evaluating positional bias may aid in areas of hallucination mitigation and position encoding optimization.

\section{Conclusion}
\label{sec:con}

In this paper, we emphasize the importance of evaluating positional bias in LVLMs. To this end, we introduce Video-LevelGauge, an extensible benchmark tailored for assessing positional bias in video understanding. We systematically evaluate 27 state-of-the-art LVLMs and reveal a significant disparity in positional bias between commercial and open-source models. 
Further analysis shows that positional bias is pervasive across different context lengths, intensifying and shifting in pattern as the context length increases. Moreover, complex contextual scenarios exacerbate the bias, while larger model variants tend to be robust. These findings suggest that future model advancement should not only focus on enhancing overall comprehension but also explicitly optimize positional bias, particularly in long video and video-text interleaved understanding tasks.

{
    \small
    \bibliographystyle{ieeenat_fullname}
    \bibliography{main}
}
\clearpage
\appendix
% \clearpage
% \setcounter{page}{1}
% \maketitlesupplementary

\section{Appendix}
\label{sec:appendix}
The Appendix is organized as follows:
\begin{itemize}
    \item Annotation Prompts (\cref{sec:prompt}).
    \item Details of Various Customized Contexts (\cref{sec:context}).
    \item Morphological Pattern Recognition (\cref{sec:MR}).
    \item More Examples of Multi-task QAs (\cref{sec:QA}).
    \item More Details of Evaluation Protocol (\cref{sec:protocol}).
    \item Effect of Probe Length (\cref{sec: porbe_len}).
\end{itemize}

\subsection{Annotation Prompts}
\label{sec:prompt}
We construct probe QAs through a three-step workflow that combines automated generation with human refinement.

\textit{(1) QA generation.} 
The prompts used for frame captioning and QA generation are presented in \cref{fig: frame_caption} and \cref{fig: qa_generation}.

\textit{(2) QA refinement.} The prompts used for for blind filtering and hallucination filtering are provided in \cref{fig: step2}. 

\textit{(3) Distractor construction.} The prompt employed to instruct LLMs in generating distractors is shown in \cref{fig: step3}. 

Task definitions are provided in \cref{fig: task_def}, and examples of discarded QAs are shown in \cref{fig: qa_dropped}.

\subsection{Details of Various Customized Contexts}
\label{sec:context}
We evaluate positional bias across six distinct context types while keeping the context length fixed. These include: (1) template context, (2) short video context, (3) long video context, (4) multi-video context, (5) long-text context, and (6) interleaved video-text context.  Configuration details are provided below.

\begin{itemize}
    \item Template context. We construct the template video by filling all frames with the ImageNet mean RGB pixel values: $[123.675, 116.28, 103.53]$.
    \item Short video context. 
    We randomly sample 100 videos from the short video subset of VideoMME~\cite{videomme}. To ensure compatibility with probe insertion, all videos are resized to match the resolution of the corresponding probes. During evaluation, each probe is randomly paired with a background video, which remains fixed throughout the positional evaluation.
    \item Long video content. 
    We randomly select 50 long videos from LVBench~\cite{lvbench}, using the same pre-processing and evaluation settings as in the short video context.
    \item Multi-video context. 
    This setting, described in Sec. 4.1, serves as our default configuration. We adopt it as an economical and effective approach to simulate long-video scenarios, especially given the substantial storage demands of natural long videos. As illustrated in Fig. 7, the long video and multi-video contexts exhibit comparable patterns of positional bias. As with the long video context, background videos are resized to match the resolution of the corresponding probe.
    \item Long-text context. 
    We extract 3,500 paragraphs from OBELICS~\cite{obelicsopen}. To ensure a context length comparable to that of video-based backgrounds, each paragraph is truncated according to the token budget of the target model. For example, for InternVL3, each paragraph is limited to approximately 1,500 tokens to match the token count of a six-frame video clip (6 frames × 256 tokens). During evaluation, nine paragraphs are randomly selected to form the background, which remains fixed throughout the positional bias assessment.
    \item Interleaved video-text context. 
    This setting is a variant of the long-text context. During evaluation, each of the nine background texts is replaced with a short video from our benchmark with a probability of 0.5, while all other settings remain unchanged.
\end{itemize}

\subsection{Morphological Pattern Recognition}
\label{sec:MR}
To recognize the morphological pattern (MR) of each model, we apply both linear and quadratic fits to the relative score (RS) at all positions. Specifically, we first perform a linear fit to estimate the global trend. Based on the slope and the mean squared error (MSE) of the fit, we clarify MR into two coarse categories: (1) monotonic type, which includes Stable with milder bias (---), Neighbor preference ($\nearrow$) and Head preference ($\searrow$) with small residuals and a relatively consistent trend; (2) non-monotonic type, which includes Lost in the middle ($\rm U$) and Volatile score ($\rm W$), typically exhibiting larger residuals that indicate deviation from linearity. We then apply quadratic fitting to the non-monotonic type to further distinguish structured non-linear patterns.

Given the relative score \{$RS_1, RS_2,\ldots,RS_N$\}, where $N$ is the number of evaluated positions, we fit the data with both linear and quadratic polynomials:
\setcounter{equation}{1}
\begin{equation}
    \hat{y}^{(1)}(x) = k x + h,
\end{equation}
\begin{equation}
    \hat{y}^{(2)}(x) = a x^2 + b x + c,
\end{equation}
where \(\hat{y}^{(1)}(x)\) and \(\hat{y}^{(2)}(x)\) denote the fitted functions from the linear and quadratic fits, respectively; \(k\) represents the slope coefficient of the linear fit.

Then we calculate the mean squared residuals for both fits separately:
\begin{equation}
    \mathrm{MSE}_1 = \frac{1}{n} \sum_{i=1}^n \left(RS_i - \hat{y}^{(1)}(x_i)\right)^2,
\end{equation}
\begin{equation}
    \mathrm{MSE}_2 = \frac{1}{n} \sum_{i=1}^n \left(RS_i - \hat{y}^{(2)}(x_i)\right)^2.
\end{equation}
where $\hat{y}^{(1)}(x_i)$ and $\hat{y}^{(2)}(x_i)$ denote the predicted value at the $i$-th position from the linear and quadratic fits, respectively.
Based on this, we categorize MR into five types:

\begin{figure*}[t]
\centering
\includegraphics[width=1.0\textwidth]{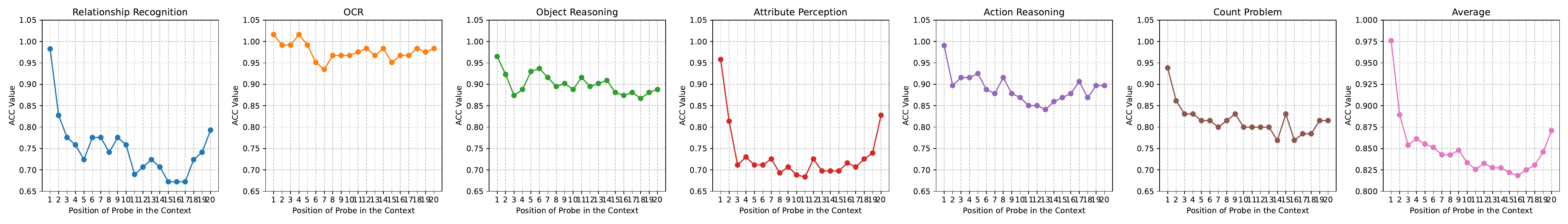} 
\caption{
Illustration of positional bias of Qwen2.5-VL-7B across six video tasks. The horizontal axis represents the probe positions within the context, while the vertical axis indicates the model’s relative score. Qwen2.5-VL exhibits reduced positional bias on the OCR task compared to its bias on other tasks.
}
\label{fig:6tasks}
\end{figure*}

\[
\text{MR} =
\begin{cases}
\text{---} & \text{if } \mathrm{MSE}_1\leq3 \text{ and } |k|\leq0.5, \\
\nearrow & \text{if } \mathrm{MSE}_1\leq3 \text{ and } k>0.5, \\
\searrow & \text{if } \mathrm{MSE}_1\leq3 \text{ and } k<-0.5, \\
\rm{U} & \text{if } \mathrm{MSE}_1>3 \text{ and } \mathrm{MSE}_2\leq2, \\
\rm{W} & \text{otherwise}.
\end{cases}
\]

\subsection{Examples of Multi-task QAs}
\label{sec:QA}
As shown in \cref{fig:6tasks}, model can exhibit varying degrees of positional bias across different tasks. Due to space constraints, only a limited number of multi-task probe examples are included in the main paper. Additional examples of multi-task QAs are provided in \cref{fig: qa_case1} and \cref{fig: qa_case2}.
Following VideoMME, we evaluate multiple-choice QA performance by prompting the LVLM with:
\begin{quote}
\textit{Select the best answer to the following multiple-choice question based on the video. Respond with only the letter (A, B, C, or D) of the correct option. QUESTION, OPTIONS. Answer with the option's letter from the given choices directly.}
\end{quote}
For open-ended items, we prompt LVLM with:
\begin{quote}
\textit{This video consists of multiple segments. You are tasked to provide a detailed description of the segment described as: QUESTION. Focus only on this part and do not describe other segments. Please describe the visual content of this segment in as detail as possible, including objects, actions, background, and any temporal progression.
}
\end{quote}
The accuracy of responses for multiple-choice QA is computed locally using exact match. For open-ended questions, following~\cite{video-chatgpt}, we use GPT-4 as an evaluator. The prompt used for scoring is provided in \cref{fig:open-ended}.

\subsection{Evaluation Protocol}
\label{sec:protocol}
We propose three statistical metrics, each capturing a distinct aspect of the model's positional bias. To enable comprehensive evaluation and facilitate comparison across models, we introduce a composite metric (CM), defined as:
\begin{equation}
    CM = \alpha_{1}P_{ran} + \alpha_{2}(100.0-P_{mean}) + \alpha_{3}P_{var}
\end{equation}
where $\alpha_{1}, \alpha_{2}, \alpha_{3}$ are balancing parameters. We empirically set them to 4.0, 3.0, and 3.0, respectively, to ensure that each component contributes comparably to the final score. Higher CM values indicate more pronounced positional bias, as depicted in \cref{fig:leaderborad}.

\subsection{Effect of Probe Length}
\label{sec: porbe_len}

Furthermore, we examine how probe length affects positional bias by varying the number of sampled frames of each probe (2, 6, and 8). As shown in \cref{tab:probe-length-ablation}, increasing probe length improves the $S_{meta}$, as richer information be sampled. Notably, positional bias basically maintains across different probe length. This suggests that a model’s robustness to positional bias is largely independent of probe length and is instead an inherent characteristic of the model.

\begin{figure}[t] 
    \centering  
    \begin{subfigure}[b]{0.23\textwidth}
        \centering
        \includegraphics[width=\textwidth]{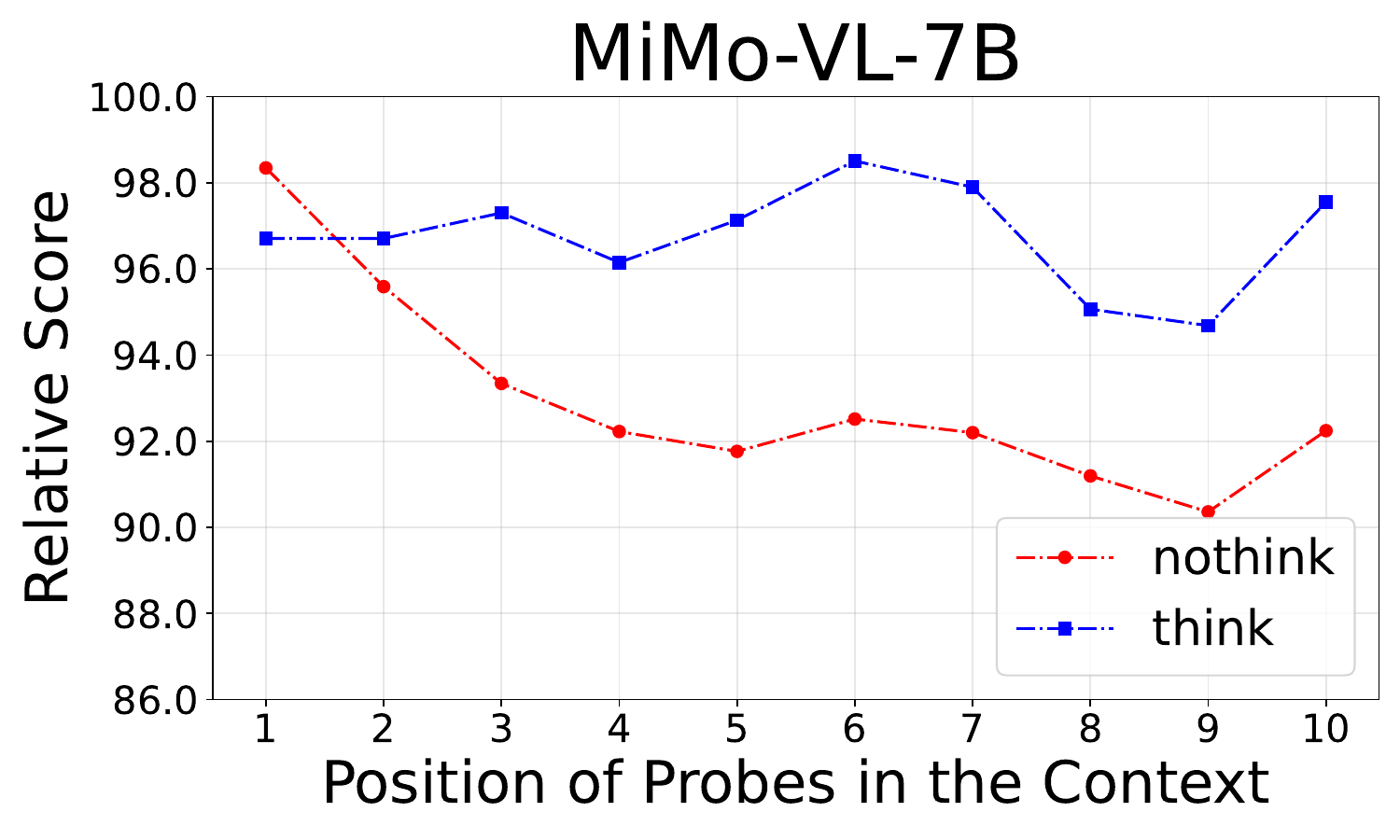} 
    \end{subfigure}
    \begin{subfigure}[b]{0.23\textwidth}
        \centering
        \includegraphics[width=\textwidth]{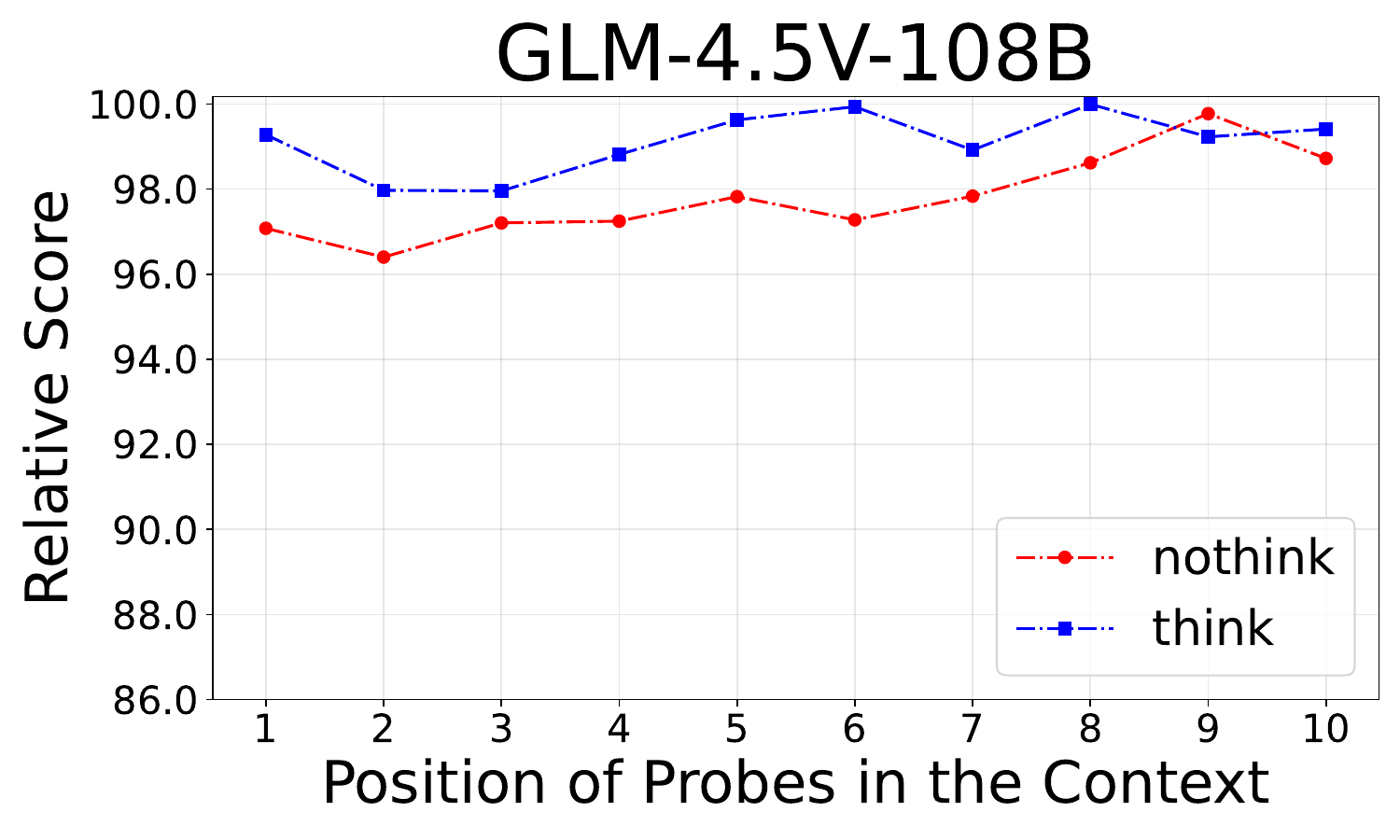} 
    \end{subfigure}
    \caption{Effect of thinking mode on positional bias. Thinking mode can alleviate the positional bias issue to a certain extent.} 
    \label{fig:thinking} 
\end{figure}

\begin{table}[t]
\centering
\caption{Effect of probe length. 
Probe length (PL) is controlled by the number of sampled frames. Although increasing the number of sampled frames significantly improves the absolute performance ($S_{meta}$), the positional bias remains largely unchanged.
}
\resizebox{1.0\columnwidth}{!}{
\begin{tabular}{lccccc}
\hline
\noalign{\vskip 2pt}
Models & PL & $P_{mean}$ & $P_{ran}$ & $P_{var}$ &$S_{meta}$ \\
\hline
\noalign{\vskip 2pt}
\multirow{3}{*}{InternVL3-8B~\cite{internvl3}} 
  & 2  &87.7&7.7&5.8&59.2 \\
  & 6  &90.3&8.0&5.3&70.5 \\
  & 8  &90.6&8.7&5.4&74.9 \\
\hline
\noalign{\vskip 2pt}
\multirow{3}{*}{LLaVA-OV-7B~\cite{llavaonevision}} 
  & 2  &88.1&8.0&5.6&54.3 \\
  & 6  &87.3&4.2&2.1&65.6 \\
  & 8  &88.2&6.5&3.2&67.7 \\
\hline
\noalign{\vskip 2pt}
\multirow{3}{*}{Qwen2.5-VL-7B~\cite{qwen25vl}}  
  & 2  &90.5&11.0&11.2&60.1\\
  & 6  &89.6&12.4&11.7&68.2\\
  & 8  &90.9&10.7&10.0&72.7\\
\hline
\end{tabular}
}
\label{tab:probe-length-ablation}
\end{table}

\begin{figure*}[t]
\centering
\includegraphics[width=0.95\textwidth]{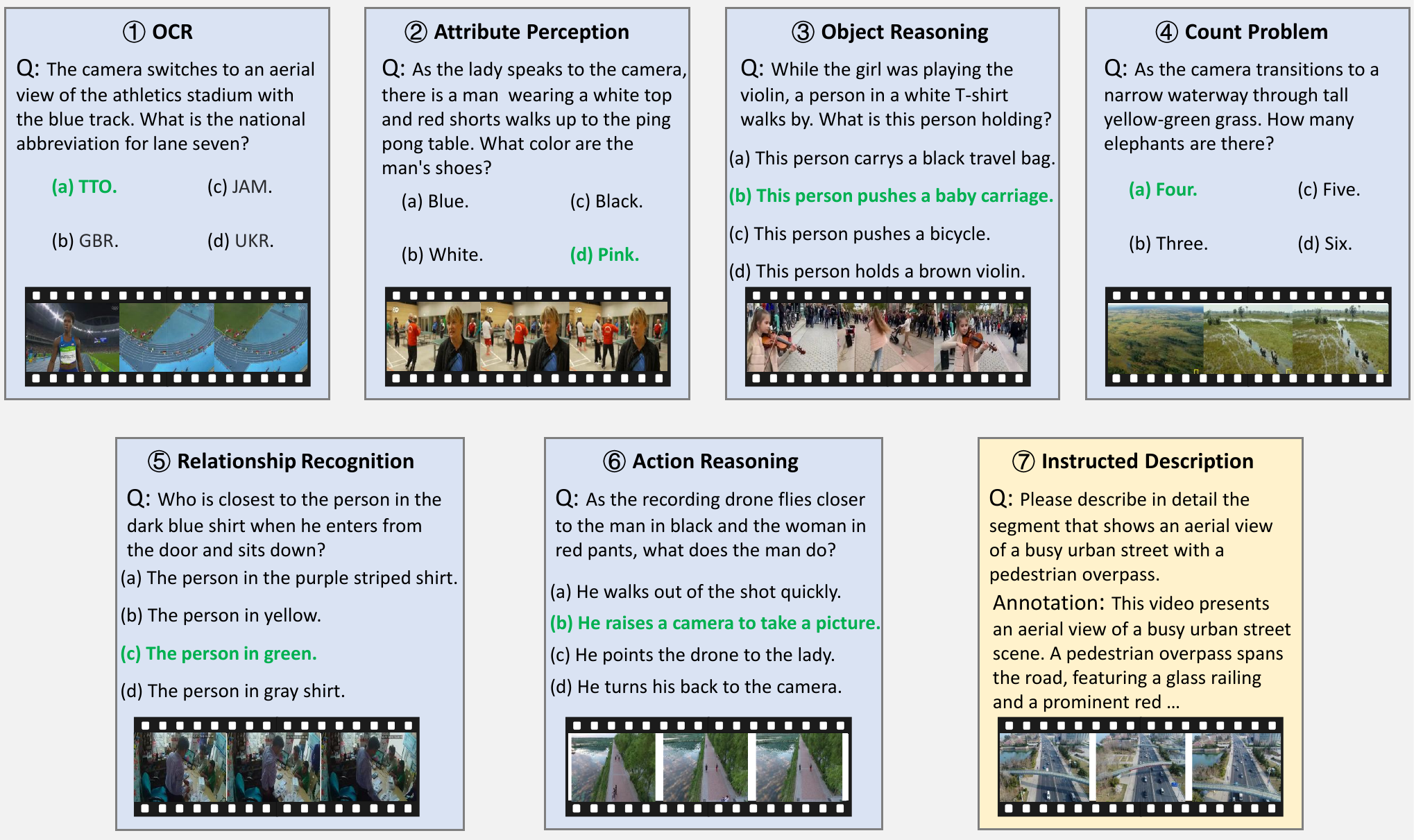} 
\caption{
Examples of probes for six MCQA formatted evaluation tasks and one open-ended instructed description task. Each question is described with scene mention and task instruction to ensure clarity, requiring genuine visual comprehension. 
}
\label{fig: qa_case1}
\end{figure*}

\begin{figure*}[t]
\centering
\includegraphics[width=0.95\textwidth]{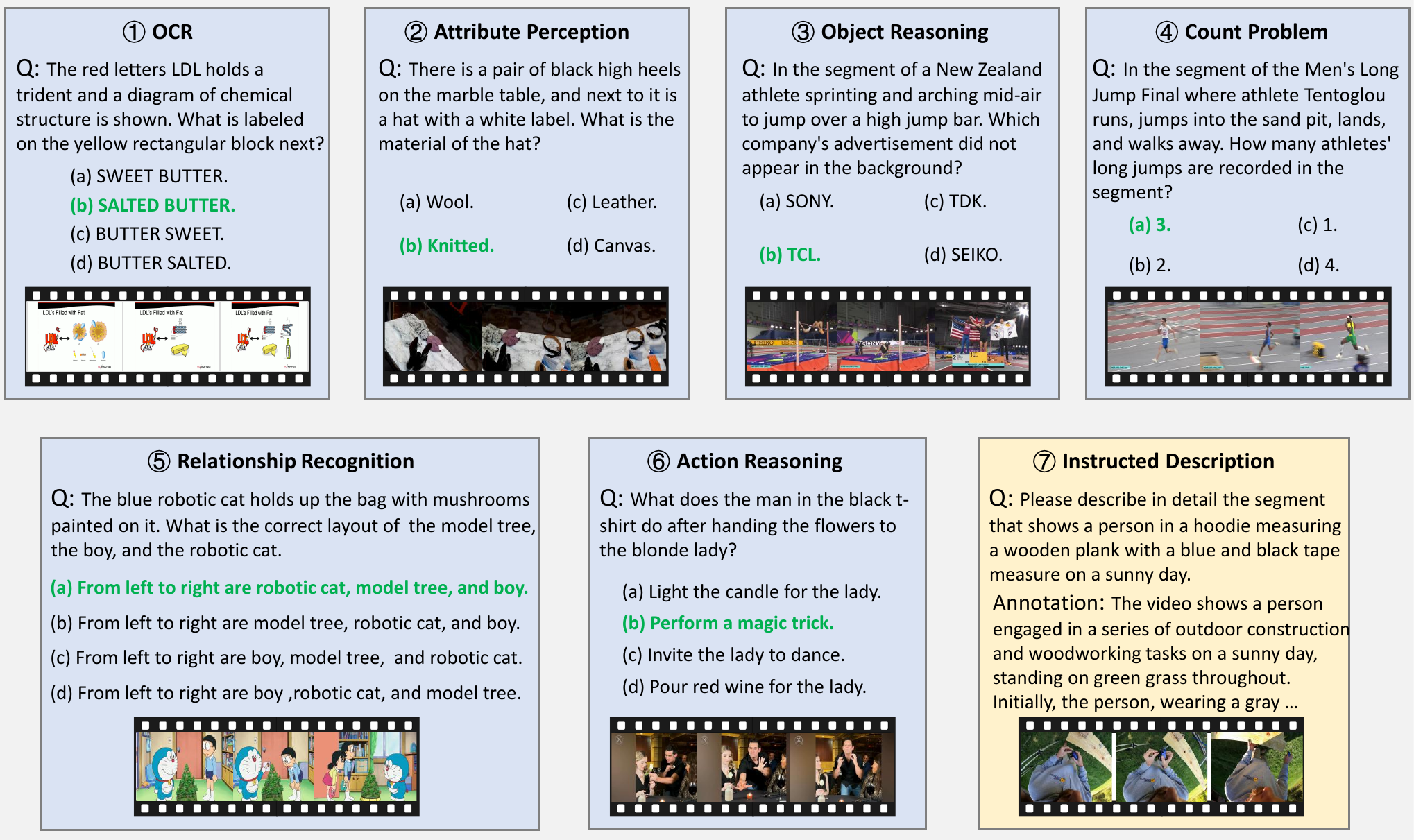} 
\caption{
Examples of probes for six MCQA formatted evaluation tasks and one open-ended instructed description task. Each question is described with scene mention and task instruction to ensure clarity, requiring genuine visual comprehension. 
}
\label{fig: qa_case2}
\end{figure*}

\begin{figure*}[t]
\centering
\includegraphics[width=0.9\textwidth]{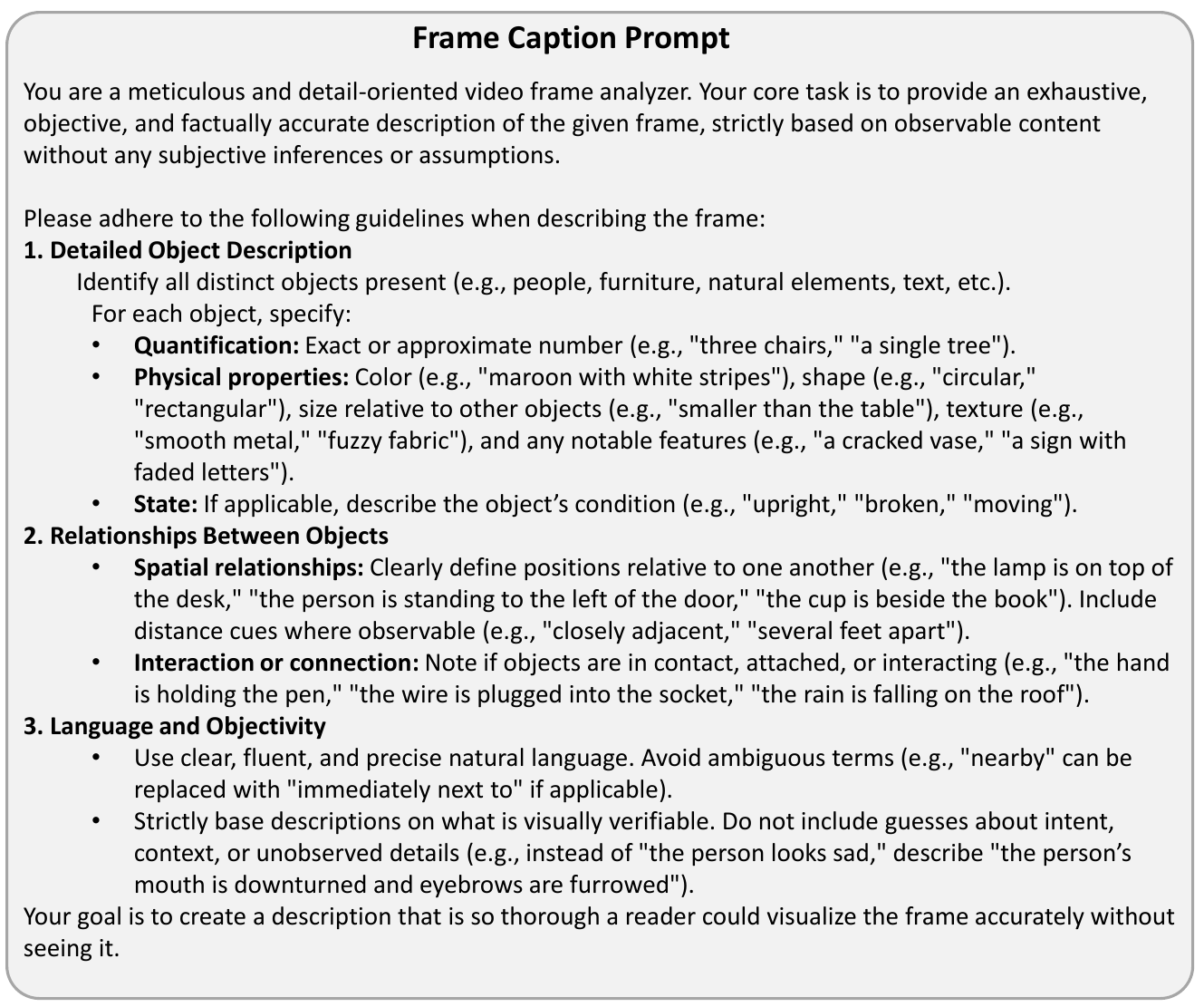} 
\caption{
The collected videos are captioned frame-wise at 1 FPS using GPT-4o with this prompt. We instruct the model to generate detailed descriptions of each frame, encompassing elements such as the number of objects, visible text, and object relationships. This detailed captioning serves as a foundation for subsequently generating multi-task question-answer pairs.
}
\label{fig: frame_caption}
\end{figure*}

\begin{figure*}[t]
\centering
\includegraphics[width=0.9\textwidth]{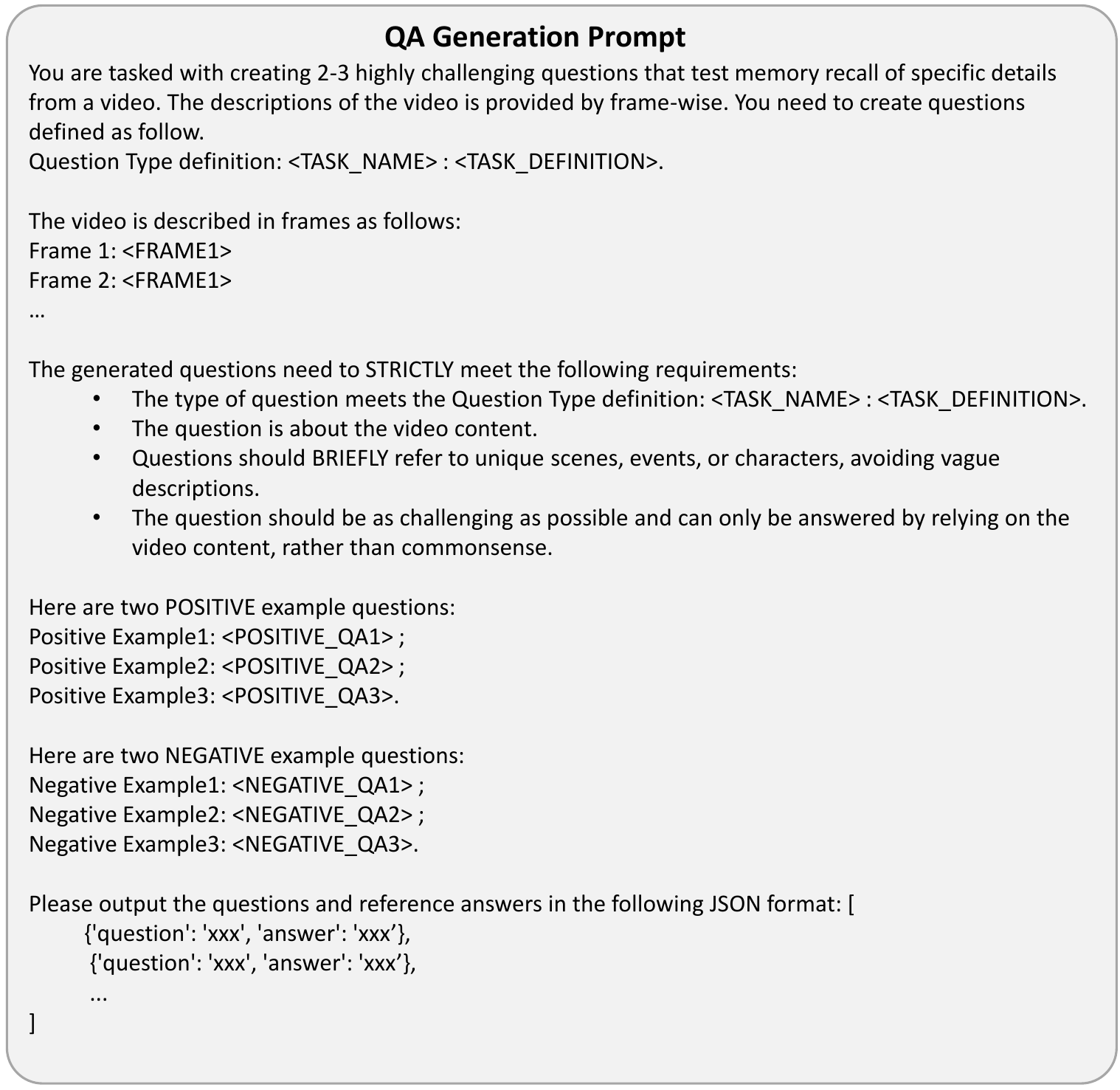} 
\caption{
Based on detailed video captions and human-annotated protocols, GPT-4 is instructed, using this prompt, to generate 2–3 task-specific question-answer pairs for each video. Different types of questions are generated separately. In total, we obtain 7,319 candidate QAs at this stage. Definitions of each task are presented in \cref{fig: task_def}.
}
\label{fig: qa_generation}
\end{figure*}

\begin{figure*}[t]
\centering
\includegraphics[width=0.85\textwidth]{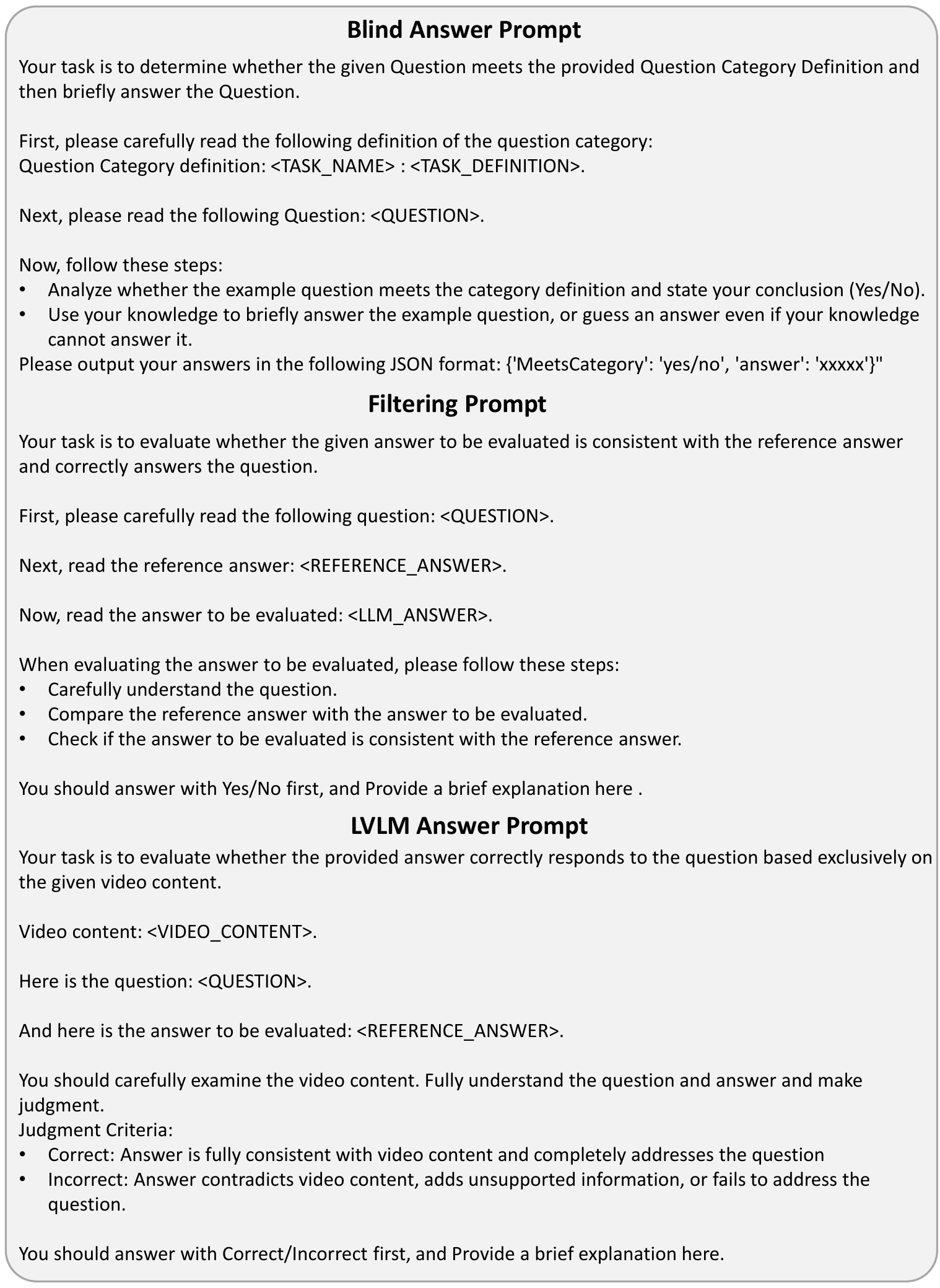} 
\caption{
During QA refinement, the generated questions are submitted to GPT-4 for blind answering using the Blind Answer Prompt. Subsequently, the QA pairs and blind answers are evaluated with the Filtering Prompt, which filters out questions answerable using text-only input. Additionally, GPT-4o is tasked with verifying the correctness of the QAs based on the original video content. Following manual selection, we obtain 1,177 final QAs and 120 descriptive items. While the construction of open-ended items has been completed, distractor generation for multiple-choice QAs requires an additional step.
}
\label{fig: step2}
\end{figure*}

\begin{figure*}[t]
\centering
\includegraphics[width=0.9\textwidth]{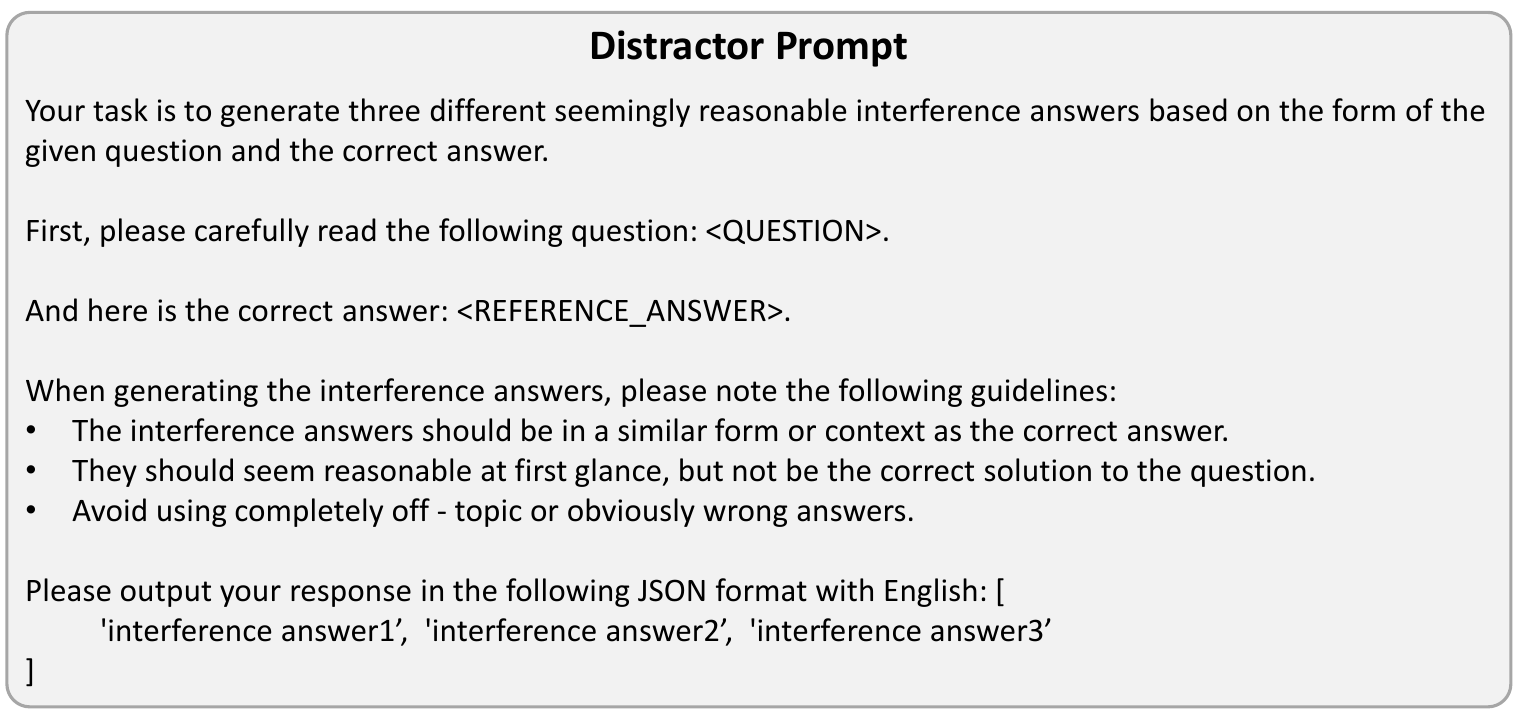} 
\caption{
For MCQAs, we input the verified question-answer pairs into GPT-4 and use a specific prompt to generate three distractors. This step is critical for creating high-quality, challenging multiple-choice questions, as low-quality distractors may lead to information leakage, for instance, enabling model to guess the correct answer easily by elimination.
}
\label{fig: step3}
\end{figure*}

\begin{figure*}[t]
\centering
\includegraphics[width=0.9\textwidth]{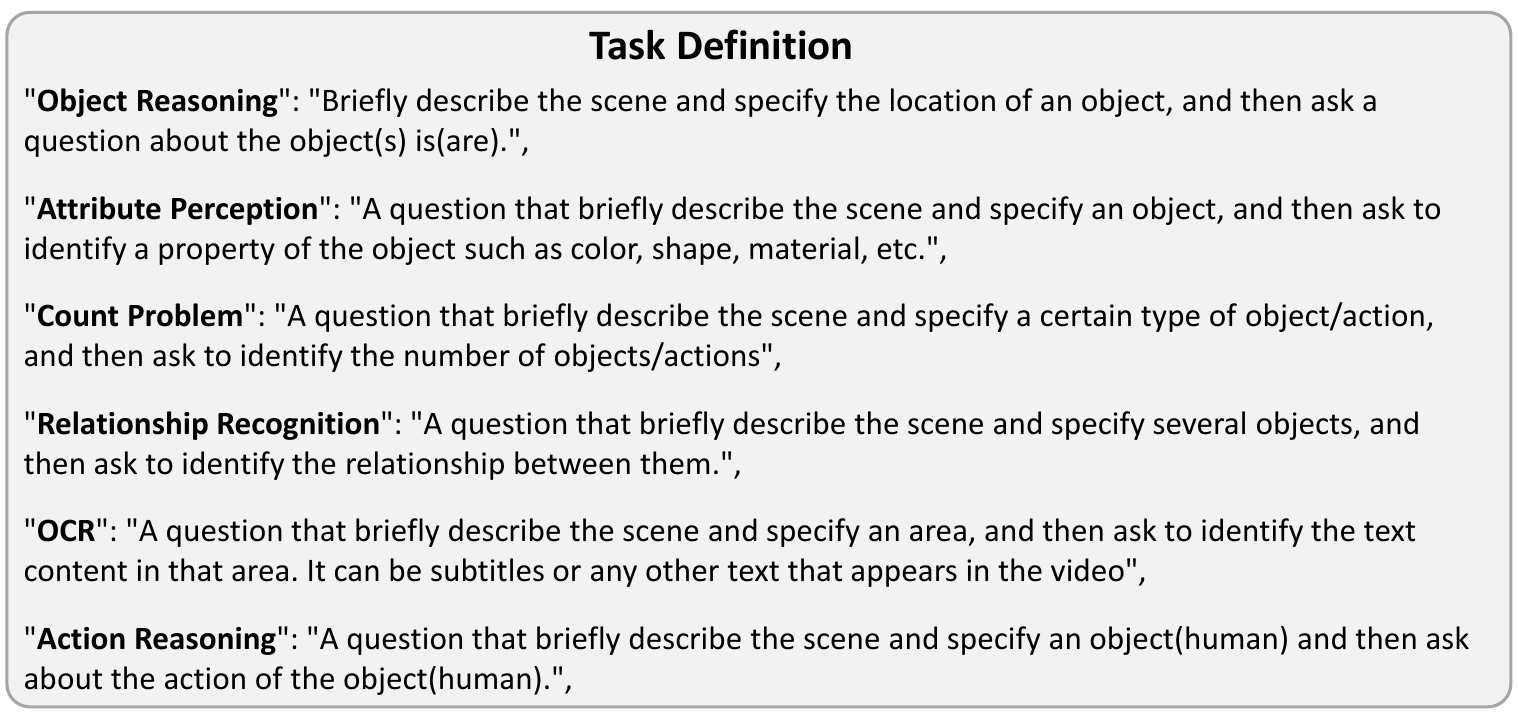} 
\caption{
Task definitions for six video understanding tasks. Additionally, descriptive questions are formulated as instructions that prompt the model to generate detailed descriptions of the video segments specified by the scene mentions.
}
\label{fig: task_def}
\end{figure*}

\begin{figure*}[t]
\centering
\includegraphics[width=0.6\textwidth]{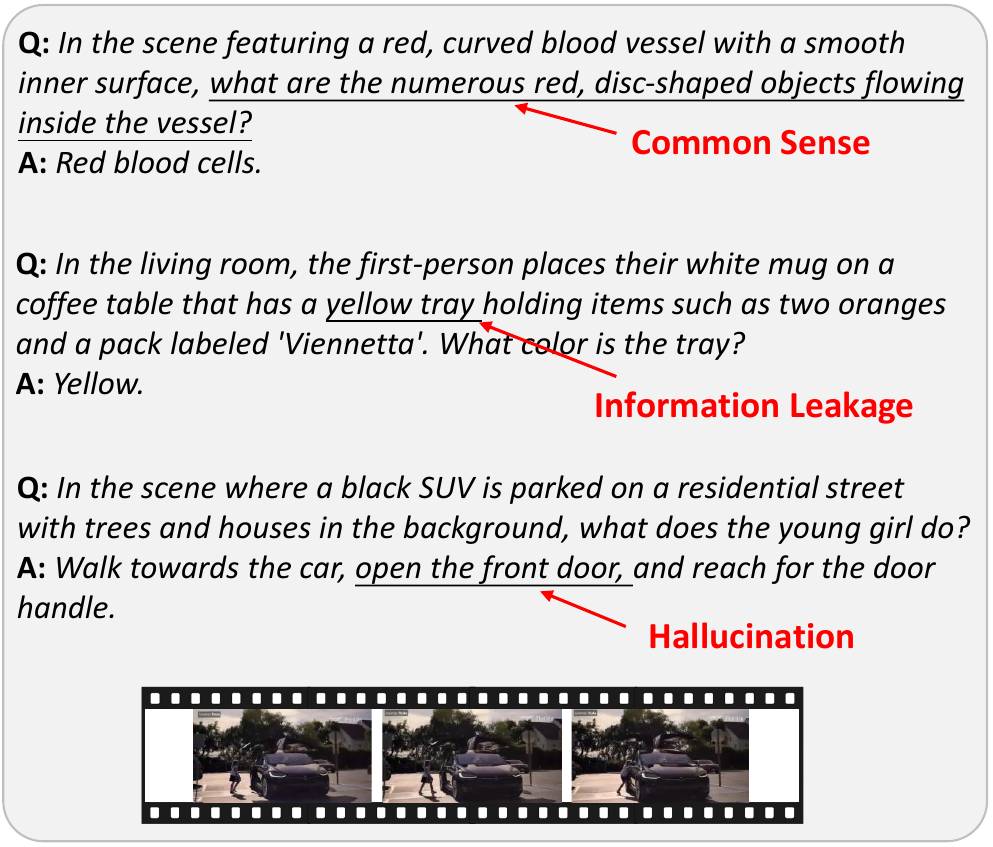} 
\caption{
Three representative examples of filtered question–answer pairs are presented. The first example illustrates a question answerable through commonsense knowledge. The second exemplifies answer leakage originating from the scene description. The third demonstrates a hallucinated response, potentially caused by inaccuracies in frame descriptions generated in Step 1.
}
\label{fig: qa_dropped}
\end{figure*}

\begin{figure*}[t]
\centering
\includegraphics[width=0.9\textwidth]{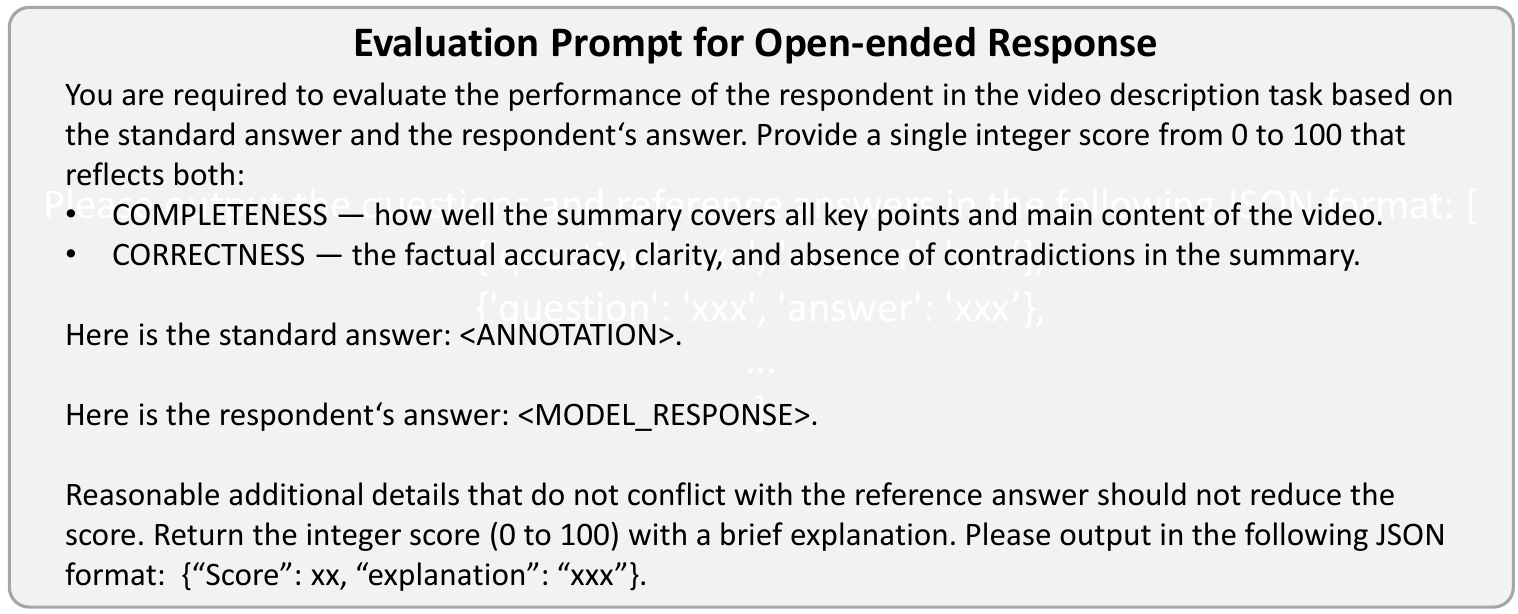} 
\caption{
We use GPT-4 to evaluate model generated open-ended outputs by comparing them against annotations.
}
\label{fig:open-ended}
\end{figure*}

\end{document}